%% file: tcp-ijcv.tex
\RequirePackage{fix-cm}

\documentclass[twocolumn]{svjour3}          
\usepackage{bm}
\smartqed  
\usepackage{times}
\usepackage{epsfig}
\usepackage{graphicx}
\usepackage{amsmath}
\usepackage{subfigure}
\usepackage{amssymb}
\usepackage[numbers]{natbib}
\usepackage{upgreek}
\include{definitions}
\include{bm}
\usepackage{color}
\usepackage{url}
\usepackage{multirow}

 \journalname{International Journal of Computer Vision}

\begin{document}

\title{Second-order Temporal Pooling for Action Recognition}


\author{Anoop Cherian \and Stephen Gould
}

\authorrunning{Second-order Temporal Pooling for Action Recognition} 
\institute{Australian Centre for Robotic Vision (ACRV), The Australian National University (ANU), Canberra, Australia,~\email{firstname.lastname@anu.edu.au}
}

\date{}

\maketitle

\input{abstract}
\input{intro}

\input{related_work}
\input{framework}

\input{expts}

\input{conclusions}

\smallskip
\noindent{\small \textbf{Acknowledgements:} This research was supported by the Australian Research Council (ARC) through the Centre of
Excellence for Robotic Vision (CE140100016) and was undertaken with the resources from the National Computational Infrastructure (NCI) at the Australian National University. The authors also thank Dr. Edison Guo (ANU) for helpful discussions. This is a pre-print of an article published in International Journal of Computer vision. The final authenticated version is available online at:\\ \url{https://doi.org/10.1007/s11263-018-1111-5}}

\bibliographystyle{plainnat}      
\bibliography{fgar_bib}  

\end{document}

%% file: definitions.tex
\newcommand{\enorm}[1]{\left\|{#1}\right\|_2}
\newcommand{\fnorm}[1]{\left\|{#1}\right\|_F}
\newcommand{\comment}[1]{}

\newcommand{\half}{\frac{1}{2}}
\newcommand{\set}[1]{\left\{#1\right\}}
\DeclareMathOperator*{\kronstack}{\bigotimes}
\newcommand{\abs}[1]{\left|{#1}\right|}

\DeclareMathOperator*{\logdet}{log \det}
\DeclareMathOperator*{\Log}{Log}
\newcommand{\reals}[1]{\mathbb{R}^{#1}}
\newcommand{\spd}[1]{\delta_{++}^{#1}}
\newcommand{\psd}[1]{\delta_{+}^{#1}}

\newcommand{\idx}[1]{[{#1}]}
\newcommand{\change}[1]{#1}
\newcommand{\newchange}[1]{#1}

\newcommand{\seq}{S}

\newcommand{\labels}{\mathcal{L}}

\DeclareMathOperator*{\MAID}{MAID}
\DeclareMathOperator*{\SMAID}{SMAID}
\newcommand{\allframesset}{\mathcal{F}}
\newcommand{\seqset}{\mathcal{S}}
\newcommand{\classifier}{p}
\DeclareMathOperator*{\cp}{TCP}
\DeclareMathOperator*{\kcp}{KCP}
\DeclareMathOperator{\bkcp}{BKCP}

\newcommand{\traj}{\bm{t}}
\newcommand{\Traj}{T}
\newcommand{\valpha}{\bm{\alpha}}
\DeclareMathOperator{\dist}{dist}
\DeclareMathOperator{\kernel}{K}
\DeclareMathOperator{\loss}{loss}

\newcommand{\mc}[1]{\multicolumn{2}{c|}{\small{#1}}}
\newcommand{\bigoh}{\mathcal{O}}


%% file: abstract.tex
\begin{abstract}
Deep learning models for video-based action recognition usually generate features for short clips (consisting of a few frames); such clip-level features are aggregated to video-level representations by computing statistics on these features. Typically zero-th (max) or the first-order (average) statistics are used. In this paper, we explore the benefits of using second-order statistics. Specifically, we propose a novel end-to-end learnable feature aggregation scheme, dubbed \emph{temporal correlation pooling} that generates an action descriptor for a video sequence by capturing the similarities between the temporal evolution of clip-level CNN features computed across the video. Such a descriptor, while being computationally cheap, also naturally encodes the co-activations of multiple CNN features, thereby providing a richer characterization of actions than their first-order counterparts. We also propose higher-order extensions of this scheme by computing correlations after embedding the CNN features in a reproducing kernel Hilbert space. We provide experiments on benchmark datasets such as HMDB-51 and UCF-101, fine-grained datasets such as MPII Cooking activities and JHMDB, as well as the recent Kinetics-600. \newchange{Our results  demonstrate the advantages of higher-order pooling schemes that when combined with hand-crafted features (as is standard practice) achieves state-of-the-art accuracy. }
\end{abstract}

\comment{
Deep classifiers for action recognition are usually trained on tiny sub-sequences. Such short temporal receptive fields may be inadequate when the actions are long, or may contain several sub-actions that needs to be inferred jointly for recognizing the action. To this end, the contributions of this paper are two-fold. First, we present~\emph{Stacked Mean of Absolute Image Differences} (SMAID), a frame-set representation that summarizes long-range spatio-temporal dynamics of video snippets by aggregating and stacking the absolute frame differences. We propose a three-stream CNN architecture using RGB, optical flow, and SMAID images for action recognition. Next, we present~\emph{Temporal Correlation Pooling}, a scheme based on the hypothesis that sub-sequence classifiers are~\emph{confused similarly for similar actions}. Using this intuition, we propose to use the correlation matrix computed on the temporal evolution of classifier scores as an action descriptor. We provide generalizations of this descriptor by embedding it into an RKHS and also approximating it via block-diagonal matrices using product quantization and model averaging when the input feature dimensionality is high. We present experiments on several benchmark action recognition datasets and demonstrate state of the art performance.
}

\comment{
Training classifiers on entire video sequences for the purpose of action recognition is often found to practically difficult due to the high computational and memory requirements. Thus, they are usually trained on individual frames or collections of them or features such as optical flow extracted from a tiny subsequence. Given that the classifiers do not have access to the full sequence, the long-range temporal aspect of the action is often lost, while the information in the sub-sequences might be shared between several actions making the classifier to be traiend uniquely for the actions. This problem becomes worse for the problem of fine-grained action recognition, which consists of actions that are subtle and difficult to differentiate from each other. In this paper, we present a scheme for representing a sequence from frame-level classifier predictions. Our scheme, dubbed~\emph{correlation pooling}, takes prediction vectors for each frame to belong to each of the action classes, the correlation matrix computed from these prediction vectors are then used to represent the sequence. These correlation matrices form Mercer kernels, thus allowing the use of RBF feature maps for embedding the prediction vectors in an infinite dimensional Hilbert space. 

As noted above, increasing the frame level information content will improve the recognition performance of the instance level classifiers. To this end, we propose a simple, but very effective image representation based on the sum of absolute image differences (SAID), which forms input to the CNN classifiers. From a fine-grained perspective, this representation amplifies the relevant human/object motions in the sequence, while also preserving motion of a short dynamic range. In comparison to the popular two-stream networks (image data and flow as separate CNN streams), our represetentation capture the joint motion and frame level representation.

We present experiments on two standard benchmark datasets. While our correlation pooling demonstrates competitive performance, combining kernelized correlation pooling with our SAID representation outperforms the state of the art on these datasets.

 The goal of this paper is to produce a sequence-level action descriptor from frame-level predictions. To this end, 
 
 Further, as noted above, increasing the temporal receptive fields of the classifiers can improve the frame-level predictions. This is usually achieved by stacking several frames or flow images; however such an approach is resource intensive. In this paper, we further present~\emph{Stacked Sum of Absolute Image Differences} (SSAID), a frame-set representation that captures the spatial and temporal dynamics of actions jointly with a longer temporal range. Compared to recent two-stream CNN models, that trains two separate CNNs on appearances and flow respectively, our joint scheme is shown to capture complimentary information not captured by either.

}

%% file: intro.tex
\section{Introduction}
\label{sec:intro}
The recent resurgence of efficient deep learning architectures has facilitated significant advances in several fundamental problems in computer vision, including human action recognition. For example, recent efforts towards action recognition using LSTM models~\cite{karpathy2014large,donahue2014long}, 3D convolutional filters~\cite{tran2014c3d,carreira2017quo}, and the two stream models and their extensions~\cite{simonyan2014two,feichtenhofer2017spatiotemporal,feichtenhofer2016convolutional} have pushed the state-of-the-art performances on standard action recognition benchmarks significantly beyond what was possible using hand-crafted features alone~\cite{wang2013dense,wang2015action}. However, despite these breakthroughs, the problem of action recognition is far from solved and continues to be challenging in a general setting. Real-world actions are often different from each other in very subtle ways (e.g., \emph{washing plates} versus \emph{washing hands}), may have strong appearance variations (e.g., \emph{slicing} cucumbers versus \emph{slicing} tomatoes), may involve significant occlusions of objects or human-body parts, may involve background activities, may use hard-to-detect objects (such as knives, peelers, etc.), and may happen over varying durations or at different rates. In this paper, we explore various second-order schemes to address some of these issues. While our schemes are applicable in a general setting, we also explore their suitability in a~\emph{fine-grained setting} that is comprised of activities having low inter-class diversity, and high intra-class diversity~\cite{rohrbach2012database}.
 \comment{
 \begin{figure}
     \centering
     \includegraphics[width=8cm,trim={0 7cm 0 1cm},clip]{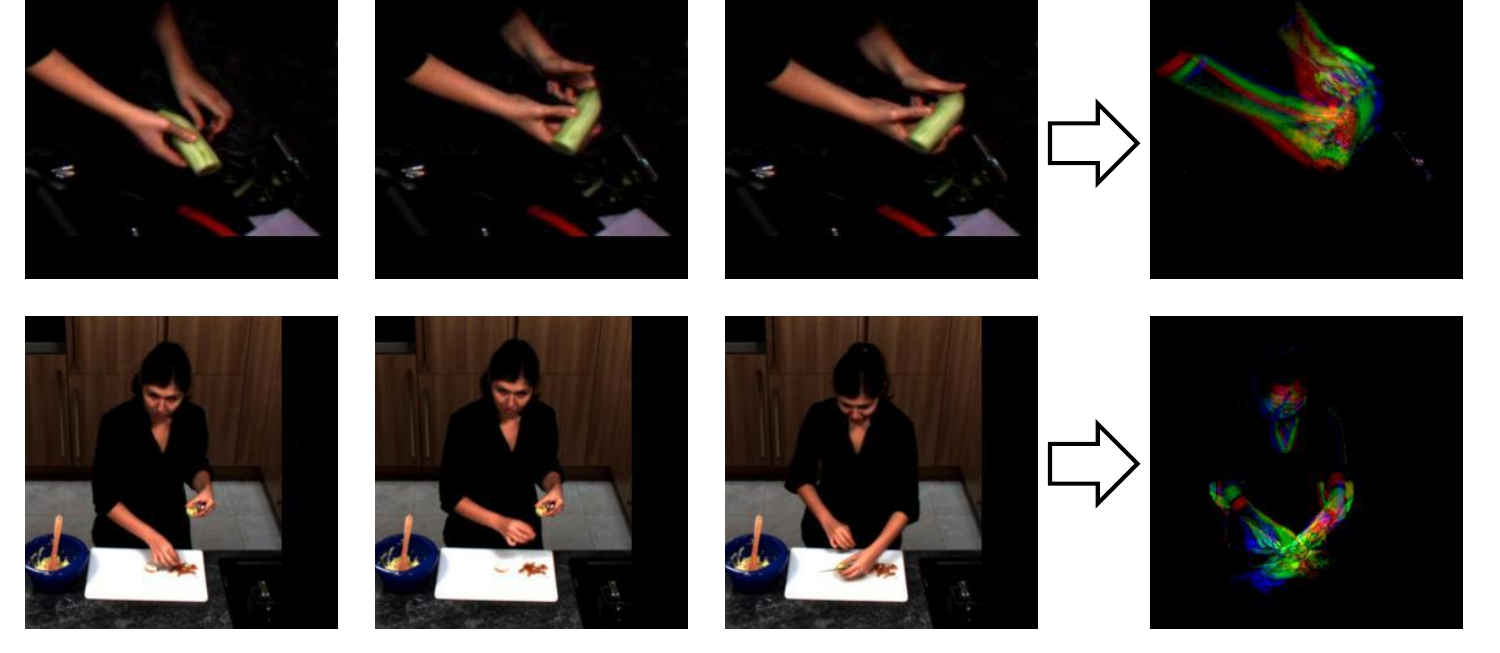}
     \caption{An illustration of our SMAID frame-set representation on sequences from the MPII cooking activities dataset. On the left are shown three frames from  sub-sequences that are used to generate a three-channel stack of MAID images shown on the right.}
     \label{fig:smaid_illustration}
 \end{figure}
 }

Most successful recent algorithms for human action recognition~\cite{simonyan2014two,feichtenhofer2016convolutional,Wang2016,srivastava2015unsupervised,yue2015beyond,carreira2017quo} are extensions of convolutional neural network (CNN) models originally designed for image-based recognition tasks~\cite{krizhevsky2012imagenet}. However, in contrast to images, video data is volumetric, and thus extending such image-based models leads to huge computational and memory overheads, which are difficult to be addressed under currently available hardware platforms. A work-around, that is often found to be promising, is to reduce the video-based recognition problem into simpler image-sized subproblems, the results from these sub-problems are later collated in a fusion layer to generate predictions for the full video. While single frames might be insufficient to capture the actions effectively as they lack any temporal aspect, using longer clips demands more CNN parameters, and thus requires more training data and computational resources. As a result, popular  deep action classifiers are trained on tiny sub-sequences (of 10--16 frames); the predictions from which are pooled to generate sequence level representations~\cite{simonyan2014two,tran2014c3d}.


Typically, max-pooling or average pooling of the sub-sequence level predictions is used~\cite{simonyan2014two,karpathy2014large,wang2015action}. Although, such pooling operations are easy to implement and fast to compute, they ignore valuable higher-level information contained in the independent predictions that could improve the recognition~\cite{cherian_grp,koniusz2016tensor,peng2016bag,hok}. For example, in the context of fine-grained recognition, let us consider two activities: ~\emph{washing plates} and ~\emph{wiping plates}. As is clear, discriminating these two actions is not easy due to their appearance similarities. Suppose sequences for the former also incorporate an overlapping activity, say~\emph{running water from tap} (which is absent in the latter). If we compute clip-level features, it is likely that some of the clips in the former will be confused between \emph{washing plates} and \emph{running water from tap}; however such a confusion is absent in \emph{wiping plates}. We propose to make use of such confusions to produce a better action representation. In the above example, we compute the co-occurrences of clip-level action classifier scores for the two activities (viz. \emph{washing plates} and \emph{wiping plates}), and then train an action classifier on these co-occurrences. As the underlying classifier confusions are strongly-correlated, the co-occurrence matrix will capture these correlations for better action discrimination, as against  using weaker statistics such as average or max pooling. 

In this paper, we propose~\emph{temporal correlation pooling} (TCP), a second-order feature pooling scheme, that takes as input a temporal sequence of CNN features (from any intermediate layer), one per video frame (Section~\ref{sec:kcp}). Each dimension of the features across time can be viewed as a~\emph{feature trajectory} corresponding to the temporal evolution of activations of the respective CNN filters. TCP summarizes these trajectories into a symmetric positive definite (SPD) matrix, each entry of this matrix capturing the similarities between such trajectories. There are several benefits that such a representation offers in contrast to prior approaches, namely (i) SPD matrices, although spanning a Euclidean subspace, are often viewed through the lens of Riemannian geometry, which offers rich non-linear distance measures for similarity computations that may help extract useful cues for recognition, (ii) SPD matrices can be naturally viewed as Mercer kernels, and similarities could be computed after embedding the feature trajectories in an infinite dimensional reproducing kernel Hilbert space (RKHS), thereby enhancing their representational power, and (iii) incorporating prior information is straightforward via sum or product kernels to the SPD kernel.

On the downside, TCP descriptors are quadratic in the size of the input features,  which may be infeasible when high-dimensional features from intermediate CNN layers are used. To circumvent this issue, we propose block-diagonal correlation matrix approximations using product quantization and model averaging. Each block matrix in the resulting representation is a small positive definite matrix and thus the above recognition framework can be directly applied. 

Another shortcoming of our pooling scheme is related to the strength of the underlying CNN model; if this model is not effective in providing reliable features, the generated descriptor will be ineffective for recognition. Although, we base our CNN on the popular two-stream model (using RGB frames for context and short stack of optical flow images for representing action dynamics), such a model is deficient in two aspects: (i) long-range temporal evolution of actions, and (ii) coupling between appearance and dynamics. While, there are several recent methods that try to address these weaknesses~\cite{yue2015beyond,Wang2016,feichtenhofer2016convolutional}, we propose a simpler workaround that is computationally very cheap, while empirically beneficial. Specifically, we propose a novel video representation dubbed \emph{Stacked Mean of Absolute Image Differences} (SMAID) that is based on averaging and stacking the absolute differences of a small set of consecutive video frames. \change{Our experiments show that SMAID captures cues complementary to appearances and optical flow, and when combined, demonstrates superior frame-level predictions, especially when the video background is stationary. } \comment{In contrast to CNN classifiers trained on single frames or flow images, our experiments show that SMAID trained classifiers demonstrate superior frame-level action predictions, especially in the context of fine-grained recognition.} Incorporating this representation, we propose a three-stream end-to-end learnable CNN framework consisting of a single frame RGB stream for action context, ten-channel optical flow stream for capturing local dynamics, and a SMAID stream capturing long-range dynamics by using subsequences, say up to 45 frames (Section~\ref{sec:smaid}).



We provide experiments (Section~\ref{sec:expts}) on four widely-used action recognition datasets to substantiate the effectiveness of our proposed schemes. \newchange{We also report results using the recent Kinetics-600 dataset~\cite{zisserman2017kinetics}, that consists of over 400K video clips, thus exploring the scalability of our approach.} Our results demonstrate that the SMAID image representation and the correlation pooling schemes demonstrate significant gains on the fine-grained task (about 4--6\%) as we expect given our motivation above. Surprisingly, they also showcase competitive performances against recent state-of-the-art methods for general action recognition. 

Before moving on, we summarize the main contributions of this paper.
\begin{itemize}
\item We propose a novel second-order pooling scheme, dubbed temporal correlation pooling (TCP)
\item We propose a kernelized variant of this pooling scheme by embedding the CNN features in an RKHS, dubbed kernelized correlation pooling (KCP)

\item We address the scalability of TCP when using higher-dimensional CNN features via our block-diagonal kernelized correlation pooling (BKCP).

\item To boost frame-level CNN predictions we propose an enhanced clip-level video representation called SMAID.

\item We propose a novel three-stream CNN action recognition model, that learns actions fusing appearance (single RGB frames), short-term (stack of optical flow), and long-term (SMAID) cues.

\item We present an end-to-end learnable variant of our CNN by providing expressions for back-propagating the gradients of a classification loss computed using TCP descriptors.

\item We provide extensive experimental comparisons on four benchmark datasets and the recently introduced Kinetics-600 dataset, demonstrating state-of-the-art performance.
\end{itemize}

%% file: related_work.tex
\section{Related Work}
\label{sec:related_work}
There is an enormous breadth of approaches aimed at tackling the problem of activity recognition. We restrict attention in this literature review to methods that have similarities to ours and refer the interested reader to recent surveys~\cite{Herath20174,chaquet2013survey} for a detailed study of this topic.
\\
\noindent\textbf{Hand-crafted Features:}~Typically, in this class of methods, features derived from spatio-temporal interest points, such as dense trajectories, HOG, SIFT, HOF, etc., are extracted from regions of interest and combined to train a discriminative classifier for action recognition. Popular methods, such as those of~\citet{wang2013dense} and~\citet{laptev2005space}, belong to this category. There have been extensions of these methods to use second-order statistical information of features via resorting to Fisher vectors (FV) in~\cite{wang2013action,sadanand2012action,oneata2013action} and stacks of FVs~\cite{peng2014action}. While we also employ higher-order statistics, we differ from these techniques in the way we encode this information. Specifically, FVs are the parameter gradients of data modeled using a Gaussian mixture models (GMM). In contrast, our method assumes the underlying CNN implicitly captures the distribution of feature vectors, and uses the empirical covariance matrix of the probabilistic evolution of classifier scores as a representation for data. Our experiments demonstrate that the proposed representation captures complementary cues to FVs, and the synergy that comes from combining our TCP encoding with FVs results in improved accuracy (Section~\ref{sec:expts}).
\\
\noindent\textbf{Deep Learning Methods:}~It is by now well-known that learning features in a data-driven way using deep learning can lead to better action representation~\cite{krizhevsky2012imagenet,simonyan2014two,ji20133d,tran2014c3d,donahue2014long,yue2015beyond}. However, as alluded to above, scarcity of annotated video data, concomitant to the demand for expensive computational resources, makes adaptability of existing machine learning algorithms to this data modality challenging; thereby demanding efficient video representations. One of the most successful of deep learning methods for action recognition is the two-stream CNN model proposed in~\citep{simonyan2014two}, which decouples the  spatial and temporal streams, thereby learning context and action dynamics separately. These streams are trained densely and independently; and at test time, their predictions are pooled. There have been extensions to this basic architecture using deeper networks and fusion of intermediate CNN layers~\cite{feichtenhofer2016convolutional,feichtenhofer2016spatiotemporal,Wang2016,wang2015action}. We also follow this trend and use a two-stream model as our baseline framework. However, we differ from these techniques in the way we use the CNN features for action recognition (first-order versus second-order). In addition, we also propose a novel three-stream CNN architecture using our SMAID image representation. 

We also note that there have been several other deep learning models devised for action modeling such as using 3D convolutional filters~\cite{tran2014c3d,carreira2017quo}, recurrent neural networks~\cite{baccouche2011sequential}, long-short term memory networks~\cite{donahue2014long,yue2015beyond}, and large scale video classification architectures~\cite{karpathy2014large}. These models demand huge collections of videos for effective training, which may be unavailable (e.g., for fine-grained activity tasks). Further, training such models with recurrent structure is also often difficult~\cite{pascanu2013difficulty}. \newchange{However, the recent emergence of very large datasets such as Kinetics-400, Kinetics-600~\cite{zisserman2017kinetics}, AVA~\cite{gu2017ava}, Moments in Time~\cite{monfort2018moments}, etc. have partially addressed the data issue. Nevertheless, state-of-art models (including 3D convolutional models~\cite{carreira2017quo}) still use clip-level feature representations that need to be aggregated via suitable pooling schemes for the final video representation or classification; thus the pooling schemes proposed in this paper are complementary to advances in CNN architectures for the action recognition problem.}
\\
\\
\noindent\textbf{Pooling Methods:}~Pooling has been an effective strategy for reducing the complexity of video representations and making them amenable to learning techniques. To this end, temporal pooling schemes have been proposed, such as 3D spatio-temporal gradients~\cite{klaser2008spatio} and STIP features~\cite{laptev2005space}. More recently, rank pooling has been proposed as an effective way for encoding the temporal evolution of actions (see, for example, \citet{Fernando_2015_CVPR,dynamic_flow,cherian_grp,cherian2018non,wang2018video}). Rank pooling, however, requires solving an order-constrained quadratic objective, which is computationally expensive. In~\citet{wang2015action}, a trajectory constrained deep feature pooling is proposed that pools features along motion trajectories. Several other CNN-based first-order temporal pooling schemes are proposed in~\cite{karpathy2014large,yue2015beyond}.

Our correlation pooling scheme is most similar to the second-order pooling approaches proposed in~\cite{carreira2012semantic,Ionescu_2015_ICCV} that also generates symmetric positive definite representations, but for the task of semantic segmentation of images. The approaches are applied on image features (such as SIFT) and cannot be easily extended to high-dimensional features generated by deep learning frameworks. In contrast, we use the frame-level prediction vectors, and the size of our correlation matrix scales by the number of action classes, which is usually much smaller than the feature dimensionality. To deal with higher dimensional features, we also propose a block-diagonal correlation matrix approximation. Our method is also different from the Riemannian geometric approaches to action recognition proposed in~\citet{guo2013action} and \citet{yuan2009human} that uses hand-crafted image features to generate covariance descriptors. 

In some earlier work~\citet{hok,koniusz2016tensor}, we briefly touch upon the idea of higher-order pooling of CNN features for action recognition, in which we explore second-order pooling as well. However, the main focus of that paper was on third-order pooling, which further requires techniques such as kernel linearization for generating descriptors of reasonable size. In contrast, in this paper we specifically explore second-order descriptors and their variants.
\\
\\
\noindent\textbf{Fine-grained Recognition:}~Early approaches  to fine-grained recognition \cite{pishchulin2014fine,rohrbach2012database,rohrbach2015recognizing} have been direct extensions of schemes described above.  Extracting mid-level appearance features, such as human body pose and motions of body-parts, have been popular for recognizing human actions \cite{prest2012weakly,wang2013approach,yao2011human,yao2012action,yao2011does,zuffi2013puppet,rohrbach2012database}). While, there have been notable advancements in human pose estimation via deep learning methods~\cite{chen2014articulated,tompson2014efficient,tompson2014joint,wei2016convolutional,newell2016stacked}, most of these models are computationally expensive and thus difficult to scale to millions of video frames that typically the datasets encompass. Moreover, most of these algorithms do not deal with occluding body-parts, which are common in long activity sequences, thus making pose-based approaches less effective. In ~\citet{cheron2015p}, human pose is used as prior to select regions of interest, and then tuning a two-stream CNN model to these regions for action recognition.
While, we do not use human pose, our SMAID representation can automatically find interesting regions with significantly less computational expense.  Other approaches to fine-grained action recognition include hierarchical multi-granularity action representations such as those depicted in~\citet{tang2012learning,lan2014hierarchical,le2011learning}, grammar based models, such as~\citet{pirsiavash2014parsing,ryoo2006recognition}, and schemes that first localize actions in a video and then detect them, such as~\citet{duchenne2009automatic,bojanowski2014weakly}. In contrast to these schemes, we use the correlations between frame-level classifier predictions to get a holistic video representation.

Another popular approach to fine-grained action recognition models human-object interactions. An object proposal framework is presented in~\citet{zhou2015interaction}, that is used to produce candidate regions containing human-object interactions, from which mid-level features based on Fisher vectors are extracted for recognizing actions. A multiscale approach is presented in~\citet{ni2014multiple} that tracks the interactions between the hand and the objects in the scene explicitly via a detection-tracking framework. A similar framework for tracking people and objects via Hough forests is proposed in~\citet{gall2011hough}. The problem has also been explored using depth cameras in~\citet{lei2012fine,wu2015watch}. While, recognizing objects is useful for recognizing actions,  frequently the objects being acted upon are occluded or might not have any discriminative features.

\noindent\textbf{SMAID Image Representation:}~The proposed video sequence summarization technique (discussed in Section~\ref{sec:smaid}) has similarities to several prior methods. Specifically, similar to SMAID, there is motion history images (MHI)~(\citet{davis1997representation}) that encodes time using image intensity (recent frames are brighter), and uses binary motion masks, thus loses texture of moving parts. SMAID uses separate image channels to capture temporal evolution. As a result, texture details of moving parts are approximately preserved per channel, while also capturing action evolution across channels. Our scheme is also different from~\citet{blank2005actions} that uses space-time volumes as shapes for recognition. More recently,~\citet{sun2014discover} and~\citet{Wang2016} also propose to use image differences as inputs for training CNN models; however they only propose a stack of single frame differences, where as SMAID uses the sum of absolute differences of several frames per channel (typically 7-10 frames), thereby capturing a longer temporal window.

\comment{
8) Action bank: A high-level representation of activity in video
10) Spatio-temporal convolutional sparse auto-encoder for sequence classification
11) Convolutional learning of spatio-temporal features
13) “Unsupervised learning of video representations using lstms
14) Trajectory based modeling of human actions with motion reference points
15) Better exploiting motion for better action recognition
17) Spatio-temporal relationship match: Video structure comparison for recognition of complex human activities
18) Action snippets: How many frames does human action recognition require?
20) Active: Activity concept transitions in video event classification.

List of papers to review:
\begin{itemize}
\item : 3D Convolutional Neural Networks for Human Action Recognition http://www.dbs.ifi.lmu.de/~yu_k/icml2010_3dcnn.pdf
\item : Sequential deep learning for Action recognition: http://link.springer.com/chapter/10.1007
\item Deep Ranking paper: very useful reference:http://users.eecs.northwestern.edu/~jwa368/pdfs/deep_ranking.pdf

\item Weakly supervised learning of interactions
between humans and objects https://www.vision.ee.ethz.ch/publications/papers/articles/eth_biwi_00828.pdf

\item Recognizing Human-Object Interactions in
Still Images by Modeling the Mutual Context
of Objects and Human Poses http://vision.stanford.edu/pdf/yaopami12.pdf

\item Learning person-object interactions for
action recognition in still images http://www.di.ens.fr/willow/pdfscurrent/delaitre_NIPS11.pdf
 
\item Fine-grained Activity Recognition with Holistic and Pose based Features http://arxiv.org/pdf/1406.1881.pdf - done
\item Exemplar-based Recognition of Human-Object Interactions 
\item What’s Cookin’? Interpreting Cooking Videos using Text, Speech and Vision http://arxiv.org/pdf/1503.01558.pdf
\item Multiple Granularity Analysis for Fine-grained Action Detection  http://www.cv-foundation.org/openaccess/content_cvpr_2014/papers/Ni_Multiple_Granularity_Analysis_2014_CVPR_paper.pdf
\item P-CNN: Pose-based CNN Features for Action Recognition  http://arxiv.org/pdf/1506.03607.pdf
\item Action Recognition by Hierarchical Mid-level Action Elements  http://arxiv.org/pdf/1508.07654.pdf
\item Recognizing Fine-Grained and Composite Activities using Hand-Centric Features and Script Data http://arxiv.org/pdf/1502.06648.pdf
\end{itemize}

}

%% file: framework.tex
\section{Proposed Scheme}
\label{sec:proposed_scheme}
We first outline our mathematical notation, followed by formally defining the activity recognition problem and our temporal correlation pooling scheme in Section~\ref{sec:tcp}. This precedes an investigation into extension of this setup for higher-order pooling (Section~\ref{sec:kcp}) and block-diagonal approximations (Section~\ref{sec:bkcp}). A brief discussion of computational complexity is given (Section \ref{sec:complexity}). We introduce our SMAID frame-set representation in Section~\ref{sec:smaid}. Next, we introduce our action classification framework using Riemannian geometry in (Section~\ref{sec:kernels}) and propose our end-to-end learnable three-stream CNN architecture (Section~\ref{sec:end2end}). 


\subsection{Notation}
We use upper-case variables (e.g., $X$) for matrices (unless defined otherwise), bold-font lower-case ($\mathbf{x}$) for vectors, and lower-case ($x$) for scalars. We use $\psd{p}$ to denote the space of $p\times p$ symmetric positive semi-definite matrices, and $\spd{p}$ to denote the same for positive definite matrices. Further, $\idx{n}$ stands for the set $\set{1,2,\ldots, n}$.

\subsection{Problem Formulation}
Let $\seqset=\set{\seq_1, \seq_2, \ldots, \seq_N}$ denote a set of $N$ video sequences, where each $\seq_i$ belongs to one of $M$ action classes with labels $\labels=\set{\ell_1,\ell_2,\ldots, \ell_M}$. Let $\seq=\left<f_1, f_2,\ldots, f_n\right>$, where each $f_i$ represents a frame, for some sequence $\seq\in\seqset$, and $\allframesset=\bigcup_{\seq\in\seqset}\set{f_i \mid f_i\in\seq}$ be the set of all frames. Our goal is to learn a function that maps any given sequence to its correct class. To this end, suppose we have trained classifiers for each action class using a training sequence set. However, we assume that it is impractical to train these classifiers on the sequences as a whole. Instead, the classifiers have been trained on individual frames. Let $\classifier_m:\allframesset\rightarrow [0,1]$ be such a classifier trained to produce a confidence score for a frame to belong to the $m$-th action class. Unfortunately, since a single frame may not be representative of the sequence, the classifier $\classifier_m$ may be inaccurate at determining the action at the sequence level. As described earlier, our goals in this paper are (i) to pool the predictions of all the classifiers from all the frames in a sequence to generate a descriptor on which sequence-level action classifiers can be trained, and (ii) to improve the confidence of each classifier $\classifier_m$ for $m = 1, 2,\ldots, M$, in making frame-level action predictions. In the sequel, we explore both these ideas.

\begin{figure*}
	\centering
	\includegraphics[width=18cm,trim={0 7cm 0 5cm},clip]{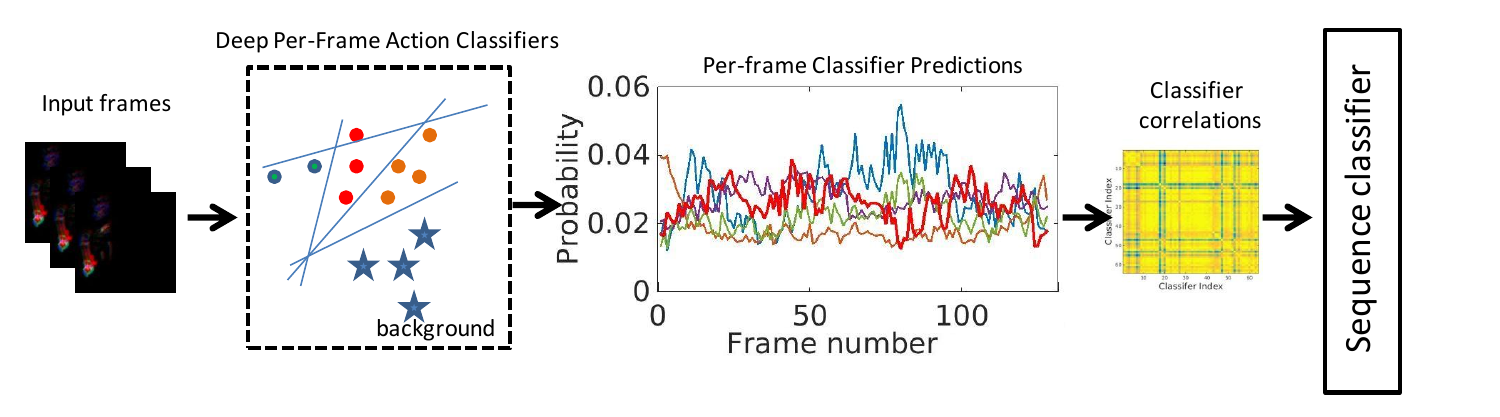}
	\caption{Internals of our correlation pooling scheme. Each video frame passes through a pre-trained set of classifiers, and their classification scores are extracted. The temporal evolution of these scores (third block above) are pooled via our correlation scheme to generate our TCP descriptor, which is then used as the action descriptor for the video.}
	\label{fig:cp_illustration}
\end{figure*}

\subsection{Temporal Correlation Pooling}
\label{sec:tcp}
Using the notation defined above, let $S=\langle f_1, f_2,\ldots, f_n\rangle$ denote a sequence of frames and let $\classifier_m(f_i)$ denote the confidence that a classifier trained for the $m$-th action class predicts $f_i$ to belong to class $\ell_m$. Further, we assume that the scores $\classifier_m(f_i)$ are normalized, so that $\sum_{m=1}^M \classifier_m(f_i) =1, \forall i\in [n]$. Let  $\valpha^m=\left(\alpha^m_1, \alpha^m_2, \ldots, \alpha^m_n\right)$ be a given vector of weights, where each $\alpha^m_i\geq 0$ and $\sum_{i=1}^n \alpha^m_i = 1$. Then
\begin{align}
\traj_m &= \left[\alpha^m_1\classifier_m(f_1), \alpha^m_2\classifier_m(f_2), \ldots, \alpha^m_n\classifier_m(f_n)\right]\\
 & \triangleq \valpha^m \circ \classifier_m(S),
\end{align}
denotes the temporal evolution of the weighted confidence of the $m$-th classifier for the frames in the sequence $S$. We call $\traj_m$ a \emph{feature trajectory}. The weights $\valpha$ give different priority to the classifier confidences across time, and is useful when there exists prior information that certain actions happen mostly at some specific regions of a sequence (e.g., beginning/middle/end).  We define our~\emph{temporal correlation pooling} action descriptor as $\cp:\reals{M\times n}\times \reals{M\times n}\rightarrow \psd{M}$, the $jk$-th entry of which is given by:
\begin{equation}
    \cp(\traj_j, \traj_k) = \sum_{i=1}^n \alpha_i^j\alpha_i^k \classifier_j(f_i) \classifier_k(f_i) = \traj_j^T\traj_{k},
    \label{eq:corr}
\end{equation}
and captures the similarity between two such feature trajectories $\traj_j$ and $\traj_{k}$ from classifiers $\classifier_j$ and $\classifier_k$, respectively. It is clear that such a similarity computes the co-activations of the classifier scores over the sequence, and thus the co-occurrences of various activities. If $\Traj\in\reals{M\times n}$ is a matrix whose $m$-th row is $\traj_m$, then taking into account the auto-correlation nature of TCP, we also define $\cp(\Traj)$ in matrix form as:
\begin{equation}
\cp(\Traj) = \Traj\Traj^T \in \psd{M},
\end{equation}
where $\psd{M}$ is the space of $M\times M$ symmetric positive semi-definite matrices. Note that, we do not center each $\traj_m$ to the mean, as is typically done when computing correlation matrices. As a result, the $m$-th diagonal entry of $\cp(\Traj)$ is given by:
\begin{equation}
\cp(\traj_m, \traj_m) = \sum_{i=1}^n \left(\alpha_i^m\right)^2\classifier_m^2(f_i) \leq \sum_{i=1}^n \alpha^m_i\classifier_m(f_i),
\end{equation}
which is the average of classifier scores, (when the $\alpha_i$'s are all set to $1/n$, this reduces to the popular \emph{average pooling} scheme). Thus, in essence the diagonal entries of TCP captures a lower bound to the first-order statistics. In the sequel, we propose to use $\cp(\Traj)$ as our action descriptor. Our full pipeline is depicted in Figure~\ref{fig:cp_illustration}.

The basic TCP scheme described above has some shortcomings: (i) it only captures second-order temporal correlations, while higher-order may be more effective, (ii) the TCP matrix will be rank-deficient if the number of frames is less than the number of action classes (which poses difficulties when using Riemannian geometric methods on them~\cite{pennec2006}), and (iii) the size of TCP is quadratic in the number of classes, thus scaling them to large feature vectors may be difficult. We address each of these issues in detail below, thereby improving the representational power of the basic TCP scheme. 

\subsection{Kernelized Correlation Pooling}
\label{sec:kcp}
From~\eqref{eq:corr}, it is easy to see that $\cp$ is a symmetric positive semi-definite matrix produced by an inner product between feature trajectories $\traj$. It is well-known that using non-linear feature maps may better capture the complex dependencies in data, leading to superior performance~\cite{vedaldi2012efficient}. To this end, we propose to embed the TCP  inner products into a reproducing kernel Hilbert space (RKHS) via the kernel trick. Mathematically, we rewrite $\cp$ in~\eqref{eq:corr} to kernelized correlation pooling ($\kcp$), where 
\begin{equation}
    \kcp(\traj_j, \traj_k) = \sum_{i=1}^n \psi\left(\alpha_i^j\classifier_j(f_i) - \alpha^k_i\classifier_k(f_i)\right), 
    \label{eq:kcp}
\end{equation}
where $\psi(x-y)$ is a suitable non-linear positive definite function. Such a reformulation brings possibilities of incorporating rich non-linearities to capture the similarities between feature trajectories. In the sequel, we use the RBF kernel
\begin{equation}
\psi(x-y) = \exp(-\gamma\enorm{x-y}^2),
\end{equation}
with a suitable choice of the bandwidth parameter $\gamma$.


\begin{figure*}[ht]
	\centering
	\includegraphics[width=14cm,trim={0 4.5cm 0 2cm},clip]{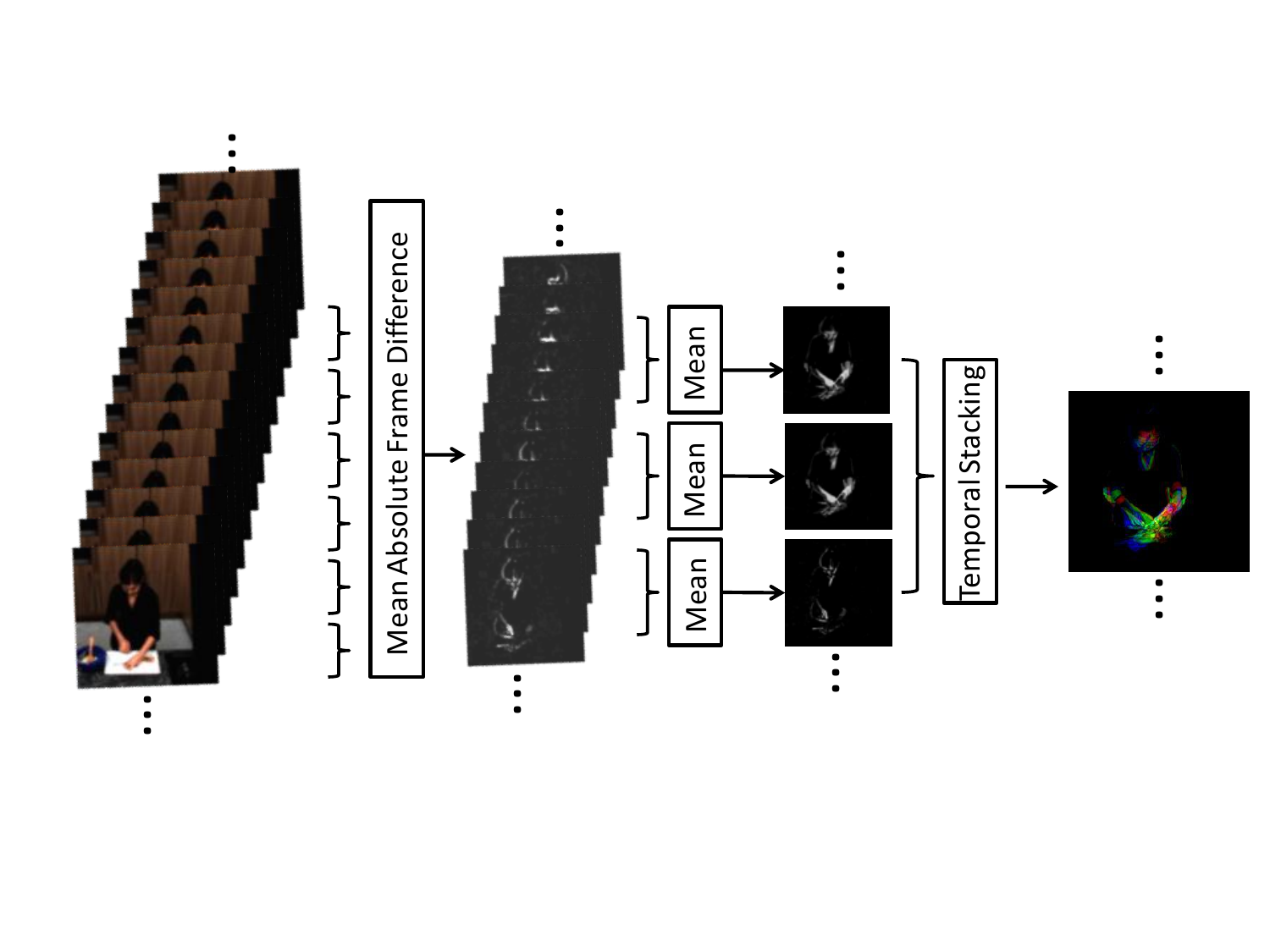}
	\caption{Schematic illustration of the steps involved in generating our SMAID images. We first convert the frames to gray-images, which are then differenced (second frame stack). These difference images are averaged to generate MAID images (third stack), which are then stacked across channels to generate SMAID.}
	\label{fig:smaid}
\end{figure*}

\subsection{Block-Diagonal Kernelized Correlation Pooling}
\label{sec:bkcp}
While, the above discussion assumed KCP is built on classifier scores, in this section, we extend it to work with any sequence of temporal features. Unfortunately, such an extension is not straightforward, because the size of KCP is quadratic in the feature size. For example, for a typical action recognition dataset, if we use the output of the last fully-connected FC8 layer (assuming a VGG/Alexnet model) and the number of action classes is 101 (used in UCF101), then KCP given a 5151-dimensional descriptor (ignoring SPD symmetry). However, extending this setup to use intermediate layer features say from FC6 or FC7 , which are 4096-dimensional, will result in KCP descriptor size of about 8 million dimensions, posing significant storage and computational difficulties. In this section, we propose a simple workaround for this problem via a KCP approximation, termed block-diagonal kernelized correlation pooling (BKCP).

In a nutshell, our main idea of the BKCP approximation is to reduce a full KCP matrix computed over all the feature dimensions into a block-diagonalized KCP, where each diagonal block of KCP captures the second-order correlations between only a subset of the features. Given that we could treat each block-diagonal of such a matrix as an independent KCP, we could scale the size of BKCP linearly in the feature size. On the downside, we ignore some correlations that could be important. To accommodate this, we repeat this BKCP construction process several times after randomly permuting the feature indexes. Such a scheme is reminiscent of the popular product quantization techniques~\cite{jegou2011product} and model averaging schemes~\cite{hoeting1999bayesian}.

Mathematically, suppose $\theta(f_i)\in\reals{d}, i\in\idx{n}$ represents features from some layer of a CNN for frames $f_i$. Further, let $\Theta\in\reals{d\times n}$ be a feature trajectory matrix built on $\theta(f_i)$ such that the $k$-th row $\Theta_k$ is given by $\theta_k(f_i), i=1,2,\cdots n$, which is the $k$-th feature trajectory. BKCP aims to quantize each $\theta(f_i)$ into the Cartesian product of several smaller features (in distinct sub-dimensions) and then compute the kernelized correlation matrix on such sub-vectors. That is, let $\theta(f_i)\in\reals{p}\times\reals{p}\times\cdots d/p\text{ times}$. Suppose, we randomly (jointly) permute the dimensions of the feature trajectories using a permutation matrix $\pi\in\Pi\subset\mathcal{P}$, where $\mathcal{P}$ is the set of all $d\times d$ permutation matrices, and we denote this shuffling as $\pi\circ\theta(f_i) = \pi(\theta)(f_i)$. Let $\pi(\theta)(f_i)^{(p(k-1)+1:kp)}\in\reals{p}$, ($k\in\idx{d/p}$), denote such a sub-vector of $p$ dimensions of $\pi(\theta)(f_i)$ starting at dimension $p(k-1)+1$. Then,  $\forall u,v\in\idx{p(k-1)+1, pk}$, we define the Block KCP (BKCP) approximation to $\kcp$ for the block at $(p(k-1)+1, p(k-1)+1)$ extending to $(pk, pk)$ in KCP as:
\begin{align}
&\bkcp^{p(k-1)+1:pk}_{\pi}(\Theta_u,\Theta_v) = \nonumber\\
&\qquad\qquad\sum_{i=1}^n\exp\left(-\gamma\enorm{\pi(\theta)_{u}(f_i) - \pi(\theta)_{v}(f_i)}^2\right),
\end{align}
where we have substituted the RBF kernel for KCP as described in the last section and the notation $\pi(\theta)_u(f_i)$ denote the $ui$-th entry of $\Theta$ after permuting its rows by $\pi$.  We extend this definition to cover all such permutations of dimensions, and we define the approximation to KCP as:
\begin{align}
&\bkcp^{p(k-1)+1:pk}(\Theta_u, \Theta_v) =\nonumber\\
&\qquad\qquad\frac{1}{|\Pi|}\sum_{\forall\pi\in\Pi}\bkcp_\pi^{p(k-1)+1:pk}(\Theta_u, \Theta_v).
\label{eq:bkcp}
\end{align}

\noindent \newchange{In words, the steps for constructing BKCP descriptors are as follows. Suppose, we use $d$-dimensional features for every frame (i.e., there are $d$ rows in the feature trajectory matrix $\Theta$). In BKCP, first we select a permutation $\pi\in\Pi$, and  permute all the rows of $\Theta$ using $\pi$. Then, we compute KCP on each disjoint set of $p$-dimensional blocks (sub-vectors or contiguous set of rows of the permuted $\Theta$). For example, the first set will have features from rows $1$ to $p$; for which we compute the KCP thereby capturing correlations between $1$ to $p$ feature dimensions, resulting in a KCP matrix of size $p\times p$. This KCP matrix is equal to some block in the matrix produced if computing KCP on all $d$ of $\Theta$; however this could be prohibitive if $d$ is large. Similarly, we compute the KCP block  for rows from $p+1$ to $2p$, and so forth. We repeat this process for all $\pi\in\Pi$. The goal of selecting different permutations is to ensure that BKCP covers a large set of inter-dimensional feature correlations of the original $d\times d$ KCP matrix. Using the above procedure, we would generate $|\Pi|d/p$ KCP blocks. Finally, these KCP blocks from different permutations are averaged to generate $d/p$ KCPs, which forms the BKCP descriptor. Specifically, if we use $|\Pi|$ permutations, then the first block of BKCP will be the average of all $|\Pi|$ KCP blocks formed from dimensions $1$ to $p$, the second BKCP block will be the average of all KCPs computed on dimensions $p+1$ to $2p$, etc.}

\newchange{This sort of BKCP construction allows creating $d/p$ KCP matrices, which is a better approximation than averaging all KCP blocks together. The latter is not a useful idea because in that case feature correlations captured in each KCP block will be lost (due to averaging a large number of KCPs), and thus performance may degrade. The same happens if we use a very large number of permutations, in which case it is straightforward to show that all the $d/p$ averaged KCP blocks will converge to the same matrix -- which is also not useful a representation. We empirically observe this effect in Figures~\ref{fig:bkcp-a},~\ref{fig:bkcp-b}, and~\ref{fig:bkcp-c}, for CNN features. Empirically, we see that more than 8 permutations will start deteriorating the action recognition accuracy.}

From an efficiency standpoint, assuming $d$-dimensional features, KCP as defined in~\eqref{eq:kcp} will have a size $d(d-1)/2$, while BKCP will have a size $d(p-1)/2$ (as each block of BKCP is symmetric)  which for appropriately chosen and fixed $p$ blocks scales linearly with $d$.  Note that we fix the permutation set $\Pi$ for all sequences in a dataset to make sure the BKCP descriptors are comparable.

\change{
\section{Computational Complexity}
\label{sec:complexity}
Using the notation defined in the previous sections, for a sequence of $n$ frames, represented by $M$ dimensional vectors, the cost of computing the TCP and the KCP descriptors is $\bigoh(M^2n)$. As for the BKCP descriptor, assuming $d$-dimensional features for every frame, $|\Pi|$ number of permutations, and using a sub-vector dimension of $p$ (then we have $d/p$ feature blocks), the cost of computing BKCP is $\bigoh(|\Pi|pdn)$. Using suitable values of $\Pi$ and $p$, the cost can be reduced significantly in comparison to finding the full $d\times d$ TCP descriptor. Note that, a na\"{i}ve compution of TCP using a $d\times n$ feature matrix costs $d^2n$ time. Choosing $\Pi$ and $p$ wisely, BKCP computations can be made significantly cheaper. For example, in our experiments, we typically use $d=4096$, $|\Pi|=8$ and $p=16$, resulting in BKCP which is 32$\times$ faster. As noted above, generating and storing the TCP descriptors for the full feature matrix is a practical concern as well. 
}
\section{SMAID Image Representations}
\label{sec:smaid}
Success of any pooling scheme depends on the quality of the features (or classifier scores) used. This is because, more noise in the features (or predictions) leads to diluting the feature correlations. While, the two stream model is popular and is empirically seen to be effective, it discards the coupling between optical flow and appearance streams. For example, in~\cite{feichtenhofer2016convolutional}, a fusion of intermediate CNN layers is proposed, where the pooling between flow and RGB streams are accounted for earlier than the last layer. Such a fusion synchronizes the two disparate feature maps and allows for joint inference at the last layer. In this section, we propose a much cheaper fusion scheme using differences of frames, that approximates flow and appearance. 

For a sub-sequence $\seq_{\tau+1:\tau+T}=\langle f_{\tau+1},\ldots, f_{\tau+\zeta}\rangle\subseteq \seq$ containing $\zeta$ consecutive frames, we define the~\emph{mean absolute image difference} (MAID) representation of $\seq_{\tau+1:\tau+\zeta}$ as:
\begin{equation}
    \MAID(\seq_{\tau+1:\tau+\zeta}) = \frac{1}{\zeta-1}\sum_{j=2}^{\zeta} \abs{f_{\tau+j}-f_{\tau+j-1}}.
    \label{eq:said}
\end{equation}
As is clear, such a representation aggregates small motions over $\zeta$ consecutive frames and summarizes them in a single object with the same dimensionality as a single frame. However, such a representation loses the long-term temporal evolution of actions; to circumvent this we stack several such MAID images corresponding to consecutive non-overlapping sub-sequences as separate image channels. That is, suppose $S' = \seq_{\tau+1:\tau+\beta\zeta}$ is a subsequence of $S$ containing $\beta \zeta$ frames. Then, we define our Stacked MAID (SMAID) representation as:
\begin{equation}
    \SMAID(S') = \kronstack_{j=1}^\beta\MAID(\seq_{\tau+(j-1)\zeta+1:\tau+j\zeta}),
    \label{eq:ssaid}
\end{equation}
\comment{
That is, suppose $S_{\tau+1:\beta\tau+\zeta} = \set{\seq_{\tau+1:\tau+\zeta}, \seq_{2\tau+1:2\tau+\zeta},\cdots, \seq_{\beta\tau+1;\beta\tau+\zeta}}$ represent $\beta$ consecutive non-overlapping sub-sequences of a sequence $\seq$. Then, we define our Stacked MAID (SMAID) representation as:
\begin{equation}
    \SMAID(\seq_{\tau+1:\beta\tau+\zeta}) = \kronstack_{j=1}^\beta\MAID(\seq_{j\tau+1:j\tau+\zeta}),
    \label{eq:ssaid}
\end{equation}
}
where the operator $\kronstack$ represents stacking $\MAID$ images into the third mode of a 3D tensor. To restrict the SMAID cross-channels to only allow temporal evolution of the actions, we reduce the original color images to gray-scale MAID images before stacking them. The overall SMAID pipeline is depicted in Figure~\ref{fig:smaid}. See  Figure~\ref{fig:qual_results} and Figure~\ref{fig:smaid_quals} for more SMAID illustrations. \change{As our representation only uses frame differences and averaging (as against, for example, the fusion scheme in~\cite{feichtenhofer2016convolutional} that needs each frame to be passed through a CNN), our scheme is computationally much cheaper. For example, differencing two frames say of size 256$\times$ 256, takes slightly less than a milli-second in Matlab on a single core desktop.}

Next, this SMAID image representation is fed to a three-stream CNN; consisting of separate streams for appearance, flow, and SMAID frames. Due to the demonstrated performance benefits, we chose a 16-layer VGG network~\cite{chatfield2014return}, pre-trained on the Imagenet dataset, to form the CNN classifiers for the individual data streams. A schematic illustration of our full pipeline is depicted in Figure~\ref{fig:cnn_architecture}.\footnote{ As we fine-tune the VGG network from a pre-trained ImageNet model, we use $\beta = 3$ for SMAID in our implementation.}


\comment{
\begin{remark}
Note that there can be alternative approaches to capturing the spatio-temporal evolution of actions  other than the one described above. For example, one could keep the two channel flow and a one channel gray scale image of appearance into a single RGB  image. However, we found that combining flow and image into one frame-stack does not help CNN train for actions. Another possibility is to use three consecutive gray scale frames, but use permutations of these frames in the frame stack such that only one arrangement of the frames is deemed to represent the positive class. However, we did not find these alternatives to lead to any benefits.
\end{remark}
}

\section{Classification on the Riemannian Manifold}
\label{sec:kernels}
\begin{figure}
    \centering
    \includegraphics[width=9cm,trim={0 4cm 0 3cm},clip]{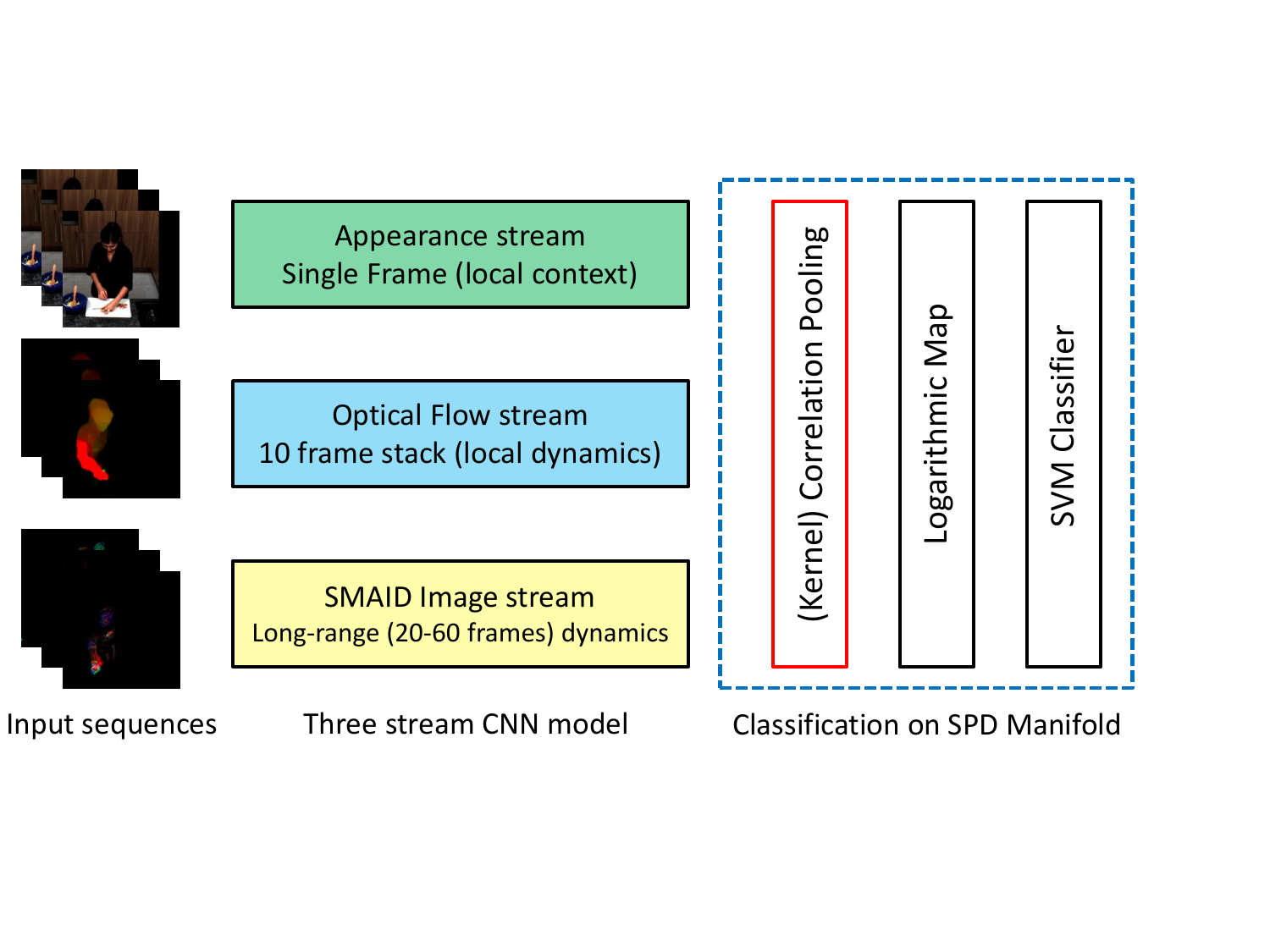}
    \caption{An illustration of our overall CNN architecture and our pooling scheme. We use a non-linear feature pooling scheme based on Riemannian geometry to generate an action descriptor.}
    \label{fig:cnn_architecture}
\end{figure}

Now that, we have provided all the details for generating a second-order action descriptor for a given video sequence, let us move on to algorithms for classifying SPD matrices in an SVM setup. Our overall classification pipeline is depicted in Figure~\ref{fig:cnn_architecture}. As is clear, the kernelized correlation matrices are symmetric positive definite (SPD) objects themselves; each sequence generating one such object. It is well-known that these matrices belong to the strict interior of the cone of positive semi-definite (PSD) matrices. While, PSD can be treated as objects in  Euclidean space under the natural Frobenius norm, it is often found that resorting to a non-linear geometry on SPD matrices can avoid unlikely or impossible outcomes (such as, for example, nearest neighbors to an SPD matrix is restricted to be only SPD matrices, instead of PSD), thereby improving application performance~\cite{pennec2006,arsigny2006log}. Typically, this non-linear geometry is imposed via the respective similarity measure used to compare SPD matrices. Among the commonly used such measures~\cite{pennec2006,cherian2011,arsigny2006log}, we will be exploring two, namely (i) the Log-Euclidean metric~\cite{arsigny2006log} and (ii) the Jensen-Bregman logdet divergence~\cite{cherian2011}, as they are known to induce valid Mercer kernels on SPD matrices. We detail each of these measures and their respective kernels  below.

\subsection{Log-Euclidean Metric}
\label{sec:log-euc}
For two KCPs $C_1, C_2\in\spd{d}$, the Log-Euclidean distance between them is given by:
\begin{equation}
\dist_{LE}(C_1, C_2) = \fnorm{\Log{C_1} - \Log{C_2}},
\end{equation}
where $\Log$ is the matrix logarithm, which makes an isomorphic mapping between an SPD matrix $C$ and a symmetric matrix $\Log{C}$, the latter uses the Euclidean geometry and thus similarity could be computed using the standard Frobenius norm. An advantage of using $\dist_{LE}$ is that it decouples the constituent matrices, such that the $\Log{}$ operator could be applied during data pre-processing, after which evaluating the similarity involves only computing Euclidean distances,  which can be done very fast. However, gradients of $\Log{}$ is an infinite series~\cite{arsigny2006log}, making end-to-end learning difficult. An RBF kernel using the Log-Euclidean metric for SVM classification is introduced in~\cite{li2013log} and has the following form:
\begin{equation}
\kernel_{LE}(C_1, C_2) = \exp\left(-\xi\dist_{LE}(C_1, C_2)^2\right),
\label{eq:log-euc} 
\end{equation}
where $\xi$ is a bandwidth parameter. Note that, the Log-Euclidean kernel can be viewed as the limit of the popular power-normalization strategy, which is known to combat~\emph{burstiness}~\cite{jegou_bursts}, i.e., certain classifiers firing more frequently than others. In addition, the Log-Euclidean kernel can be directly applied to each block of our BKCP descriptor separately, thus making the scheme efficient (as otherwise one needs to compute the singular values of a very large KCP matrix).
  
\subsection{Jensen-Bregman Log-Det Divergence}
Another popular similarity measure on SPD matrices is the recent Jensen-Bregman Log-Det divergence (JBLD)~\cite{cherian2011} (also called Stein divergence~\cite{sra2011positive}), which for two KCPs $C_1$ and $C_2$ has the following form:
\begin{equation}
\dist_{S}(C_1, C_2) = \logdet\left(\frac{C_1+C_2}{2}\right) - \half\logdet{\left(C_1 C_2\right)}.
\label{eq:jbld}
\end{equation}
In contrast to the Log-Euclidean metric, JBLD is not a Riemmannian measure, instead is a symmetric Bregman divergence which captures the information divergence between a function and its first-order Taylor approximation (the function is $-\logdet$ in this case). It is related to the Bhattacharya distance~\cite{jebara2003bhattacharyya} between two zero mean Gaussian distributions with covariances $C_1$ and $C_2$. In contrast to the Log-Euclidean metric that needs to compute the matrix logarithm of the constituent matrices, JBLD needs only the matrix determinant, which is computationally cheaper. In~\cite{sra2011positive}, a kernel is defined using JBLD as defined below:
\begin{align}
    \kernel_{S}(C_1, C_2) &= \exp\left(-\xi \dist_{S}(C_1, C_2)\right),\\
    \forall\xi &\in \set{\frac{k}{2}, k=1,\cdots,d-1}\cup \left[d,\infty\right),\nonumber
\end{align}
where the bandwidth parameter $\xi$ is defined only for certain values. In contrast to the Log-Euclidean metric, JBLD offers computationally cheaper gradients, as will be explored in the next section.

\section{End-to-End CNN Training}
\label{sec:end2end}
In this section, we explore an end-to-end CNN architecture that learns the action descriptors and the classifiers jointly via gradient back-propagation. As is the case with any end-to-end CNN models, the main challenge in designing this model is to define the gradients of the objective with respect to the inputs. There have been several previous attempts at implementing end-to-end second-order CNN models. In~\cite{Ionescu_2015_ICCV}, the Log-Euclidean metric is used to define the CNN loss function. While computing gradients of this metric is challenging (as $\Log{}$ involves an SVD operation which by itself is expensive when it needs to be done a large number of times within optimization schemes such as stochastic gradient descent), it also demands flattening of the matrix, leading to very large fully-connected layers that scales quadratically with the number of data classes. In~\citet{huang2016riemannian}, a CNN model that takes SPD matrices as input is presented. Another recent attempt (~\citet{yu2017second}) is to map the second-order SPD matrices into a lower-dimensional SPD manifold through parametric second-order transformation, followed by parametric vectorization. However, such parametric transforms also introduce additional capacity to the networks that needs to be learned. In contrast to all these methods, we propose to directly use second-order similarity measures to define loss functions, which as we show below leads to simple and straightforward gradient formulations, without the need for introducing any new parameters into the framework. We explore two such loss functions, namely (i) using the Jensen-Bregman Logdet Divergence as introduced in~\eqref{eq:jbld}, and (ii) using the simple Frobenius norm.

\subsection{End-to-End Learning Using Stein Divergence}
Suppose $T\in\reals{M\times n}$ denotes the CNN feature trajectories\footnote{With a slight abuse of previously introduced notations, we assume $T$ to be raw feature trajectories without any scaling or normalization.} (from say the FC8 layer of a standard VGG/ Alexnet model) for $n$ frames in a sequence and $M$ action classes. Further, let $Y$ denote an $M\times M$ diagonal ground-truth label matrix for a ground-truth label $\ell$ associated with $T$; the $jj$-th diagonal entry of $Y$ is defined as
\begin{equation}
Y_{jj} =\begin{cases} 1/\left(1+(M-1)\epsilon\right),\qquad \text{if } j=\ell_i\\
		\epsilon/\left(1+(M-1)\epsilon\right),\qquad \forall j\neq \ell_i
        \end{cases}
\label{eq:label_matrix}
\end{equation}
where we assume $\epsilon$ is a small number (say $10^{-5}$ used in our experiments). This encoding of ground truth class label is similar to the standard one-hot encoding used with a softmax cross-entropy loss framework. However, given that we propose to use similarity measures defined on SPD matrices in our loss, we cast the label in a matrix form and use a small $\epsilon$ regularization to make sure this matrix SPD.

Suppose, we have a training set consisting of such sequences of CNN feature trajectories $\mathcal{T} = \set{T^1, T^2,\cdots T^N}$ for video sequences in $\seqset$ and their associated ground-truth encoded matrices $\mathcal{Y} = \set{Y^1, Y^2, \cdots, Y^N}$. Then using the JBLD measure introduced in~\eqref{eq:jbld}, we define the TCP CNN loss as:
\begin{align}
& \loss(\mathcal{T}, \mathcal{Y}) := \sum_{\forall (T,Y)\in \mathcal{T}\times\mathcal{Y}} \bigg[ \log\det{\left(\frac{Y + \cp(T)}{2}\right)}\nonumber\\ 
&\qquad\qquad\quad-\half\log\det{Y} -\half \log\det{\left(\cp(T)\right)}\bigg],\\
&\text{\quad where } \cp(T) = \frac{1}{n} T{T}^T. \nonumber
\end{align}

For implementing back-propagation, we need the gradient of $\loss$ with respect to a data matrix $T$ (with associated label matrix $Y$) and is as follows:
\begin{equation}
\frac{\partial \loss(T, Y)}{\partial T}=\frac{2}{n}\left\{\bigg(\cp(T)\!+\!Y\bigg)^{-1}\!\!\!\!\!\!\!-\half\cp(T)^{-1}\right\} T.  
\label{eq:jbldgrad}
\end{equation}

\subsection{End-to-End Learning Using Frobenius Norm}
A difficulty usually encountered with the gradient defined in~\eqref{eq:jbldgrad} is the need to compute the matrix inverse, which is expensive and will also sometimes lead to numerical instability. Thus, we also propose to use the matrix Frobenius norm to define the CNN loss, which avoids these issues. As this loss will not require the label matrix $Y$ to be SPD, we assume $\epsilon=0$ in this case in~\eqref{eq:label_matrix}. Reusing the notations from the last section, we define the new loss as:
\begin{align}
& \loss(\mathcal{T}, \mathcal{Y}) := \sum_{\forall (T,Y)\in \mathcal{T}\times\mathcal{Y}} \fnorm{ \cp(T) - Y }^2, 
\end{align}
and the respective gradient with respect to a data matrix $T$ has the form:
\begin{equation}
\frac{\partial \loss(T, Y)}{\partial T} =\frac{2}{n}\bigg(\cp(T) - Y\bigg) T  
\label{sec:frobgrad}
\end{equation}

Empirically, it is observed that using the softmax output of the FC8 CNN layer for constructing the above losses leads to better convergence of the models.

\comment{
For implementing back-propagation, we need to define two gradients, namely (i) the gradient of the loss and (ii) the gradient of $\cp$, both with respect to $T$. Below, we provide the respective expressions. The gradient of the loss with respect to $T$ can be reduced to that with respect to the JBLD measure and is:
\begin{align}
&\frac{\partial \dist_{S}(\cp(T), Y)}{\partial T} = \nonumber\\
	&\quad\left[\left(\cp(T)+ Y\right)^{-1} - \half \left(\cp(T)\right)^{-1}\right] \times \frac{\partial \cp(T)}{\partial T},    
\end{align}
where the $jk$-th element of the TCP gradient is given by:
\begin{equation}
\frac{\partial \cp(T)}{\partial T_{jk}} = \frac{1}{n}\left[T I_{kj} + I_{jk} {T}^T\right],
\end{equation}
where $I_{jk}$ is a $M\times d$ binary matrix with all zeros except a one at the $jk$-th entry.
}
\comment{
Such frame-level predictions are later pooled using average or max-pooling to produce a label for the full video sequence. Such a scheme might be disadvantageous to a fine-grained setting because (i) the frame-level predictions work best when there are discriminative frames in the sequence that can unequivocally predict the action class, and (ii) separated learning of the appearance and flow can lose the otherwise important coupling between them. The problem (i) might not be significant in a general action recognition setting where there are distinct motion patterns or appearances to signify the action (examples include walking, running, throwing, etc.), the problem becomes challenging in a fine-grained setting where there might not be sufficient frame-level cues for discriminating between actions (such as, say, slicing cucumbers against vegetables); perhaps more context can help improve the frame-level predictions. For (ii), there have been extensions of the frame-level predictor to use more context by stacking several frames or flow images there by capturing a greater temporal receptive field. Such implementations are memory intensive and slow in training as the number of CNN parameters will also increase.}

%% file: expts.tex
\section{Experiments}
\label{sec:expts}
In this section, we evaluate the usefulness of our proposed framework on four datasets. Two of these datasets, namely  the MPII Cooking activities dataset~\cite{rohrbach2012database}, and the JHMDB dataset~\cite{jhuang2013towards}, are standard fine-grained benchmarks. We also provide evaluations on HMDB and UCF101 datasets, which are standard benchmarks with fine-grained as well as coarse action categories. \change{As for the CNN architecture, we report results using Alexnet~\cite{krizhevsky2012imagenet}, VGG-16~\cite{chatfield2014return}, and ResNet-152~\cite{He_2016_CVPR}, demonstrating that the benefits showcased by our representations are CNN architecture agnostic.} Below, we provide details of these datasets, data preparations, evaluation protocols, and our results. Later, in Section~\ref{sec:kinetics}, we provide experimental results on the large-scale Kinetics-600 dataset.

\subsection{Datasets}
\paragraph{MPII Cooking Activities Dataset~\cite{rohrbach2012database}:} This dataset consists of high-resolution videos of cooking activities captured by a static camera. The videos are of 14 different people cooking various dishes and consists of 64 distinct activities spread across 3748 video clips and one background activity (1861 clips). There are over 800K frames and the activities range from coarse subject motions such as~\emph{moving from X to Y},~\emph{opening refrigerator}, etc., to fine-grained actions such as \emph{peel},~\emph{slice}, \emph{washing hands}, \emph{cut ends}, \emph{cut apart}, etc. This dataset is challenging due to several reasons, namely (i) the classes are very unbalanced -- there are certain activities that have only about 1K frames over the entire dataset, (ii) there is significant intra-class variability as the participants are only asked to prepare one of a set of 14 dishes and allowed to cook in their own styles, and (iii) there are no annotations of objects in the scene, and the tools used for actions are very small (such as spice folder, knife, etc.) and thus hard to detect. 

\paragraph{HMDB Dataset~\cite{kuehne2011hmdb}:} It consists of 6766 videos from 51 different action categories, mostly web videos of low resolution and quality. Each video clip is a few seconds long. The actions in these clips vary significantly in lighting, and viewpoints, and may have significant camera motions making the action recognition task challenging. The dataset includes videos that are not person centered and the actor may undergo occlusions as well.

\paragraph{JHMDB Dataset~\cite{jhuang2013towards}:} This dataset is a subset (960 videos) of the HMDB dataset consisting of 21 actions, but contains videos for which the human limbs can be clearly identified. It was primarily designed for action recognition using human poses. The dataset contains action categories such as \emph{brush hair}, ~\emph{pick},~\emph{pour},~\emph{push}, etc. 

\paragraph{UCF101 Dataset~\cite{soomro2012ucf101}:} This dataset contains 13320 videos spread in 101 action categories. The dataset is different from the above ones in that in addition to several of the categories found in HMDB dataset, it also contains videos on sports activities; such videos usually have strong camera motions, long shots (and thereby person occupying very small portions of the scene), and fast actions. The clips in this dataset are also of low resolution and of web quality. A few illustrative actions in this dataset are \emph{cartwheel}, \emph{somersault}, \emph{kayaking}, \emph{Tennis swing}, etc. and also includes non-sports actions such as \emph{apply eye makeup}, \emph{brushing teeth}, etc., similar to the ones in the HMDB dataset.

\paragraph{Evaluation Protocols:} Following the standard protocols, we use mean average precision over 7-fold cross-validation on the MPII dataset. Other datasets use mean average accuracy on 3-splits. For the former, we use the evaluation code published with the dataset.
\begin{figure*}
\centering
\includegraphics[width=17cm,trim={0cm 5cm 0cm 4.5cm}, clip]{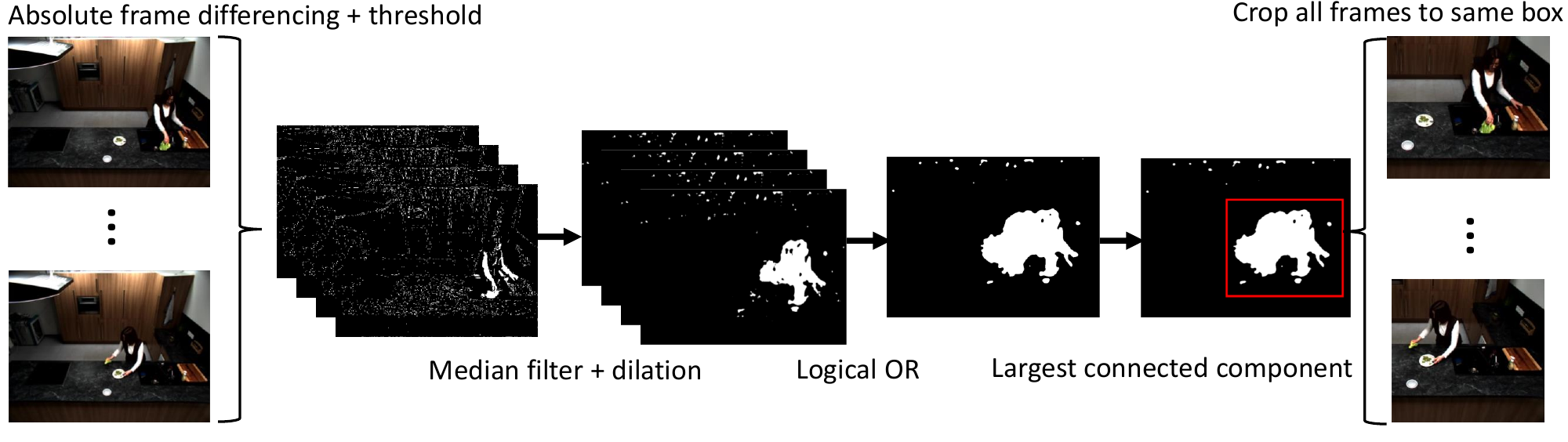}
\caption{\newchange{Preprocessing pipeline for the MPII dataset. Our goal is to generate the smallest bounding box (for the full clip) that will contain the moving parts in the video; subsequently cropping the high-resolution frames to this box. As the crop box is computed for all frames, the background is kept constant after the crop, while the relevant action dynamics are contained in this box.}}
\label{fig:preprocess_mpii}
\end{figure*}
\subsection{Preprocessing} \change{The original MPII cooking videos are very high resolution (1624$\times$1224), however the actions happen only at certain parts of the scene. Given that such full resolution frames cannot be directly used to train the CNNs, and resizing the frames to a CNN input resolution might reduce the number of pixels belonging to the actions, an attention mechanism is important to crop the frames around regions around actions. Further, we would also want to use the fact that the camera is static (which will be useful to compute SMAID images). Thus, we use morphological operations to compute these action regions, as detailed below. We found that using a person detector (such as using a faster-RCNN~\cite{ren2015faster}) per frame -- that returns a person bounding box, and then finding a crop box for the sequence that is a rectangular hull of all the frame level boxes -- will lead to similar results.}

As alluded to above, instead of using a faster RCNN for finding the crop box that needs every frame to be passed through a CNN, we resort to a simple set of morphological operations, that are computationally much cheaper and produces the same result. \newchange{The preprocessing pipeline is illustrated in Figure~\ref{fig:preprocess_mpii}.} Specifically, for every sequence, we first convert the frames to half their sizes, followed by frame-differencing that produces appearance blobs corresponding to the moving parts (mostly parts of human body, such as hands). We then dilate these blobs using a $11\times 11$ dilation filter to capture details surrounding them. This is followed by Gaussian smoothing and connected components analysis to find blobs connected to each other. The connected components are converted to binary masks, and are merged with such components across frames in the sequence (logical OR). For example, in the case when a person moves from say X to Y in the sequence, our scheme results in a binary mask of the person per frame, and such masks are merged (as they will be connected due to the neighborhood) across frames, resulting in one large blob for the motion from X to Y. We then use the largest such merged binary blob and crop the sequence to a box containing this blob. The cropped frames are then resized such that their shorter side is 256 pixels, to be used for training the CNNs. We use these resized frames for computing optical flow using the TVL1 OpenCV implementation. Each flow image is then thresholded to $\pm$ 20 pixels, rescaled to 0--255, and saved as a JPEG image for storage efficiency as described in~\cite{simonyan2014two}.

\change{For the JHMDB dataset, we use the RGB frames resized such that the shorter side is 256 pixels, and compute optical flow on them directly using the same scheme described above. For the UCF101 and HMDB datasets, we use the pre-processed frames and flow images publicly shared as part of two-stream fusion implementation\footnote{Available from \url{https://github.com/feichtenhofer/twostreamfusion}}~\cite{feichtenhofer2016convolutional}.}


\subsection{Experiment Setup} All the three CNN streams (RGB, Flow, and SMAID) are trained separately. Among the end-to-end CNN loss variants (Frobenius norm versus Stein divergence), we use the  Frobenius norm due to its superior speed and numerical stability. We found that the performance of Frobenius norm is very similar to the standard softmax cross-entropy loss. We use sub-sequences of 30 frames  for computing the correlation matrices in the end-to-end setup. Given a fixed CNN batch size (number of frames), we could not use more frames per sequence, as this limits the number of sequences that could be used in a training batch, and thus restricting the batch diversity (different action classes in the same batch). Less diverse such batches are known to impact convergence. Once the CNNs are trained, we use a forward pass to compute per-frame features, which is then used to generate sequence level TCP descriptors and variants. These descriptors are then used in a Riemannian geometry based SVM classification framework, thus utilizing the power of non-linear geometry. We found that this provided significantly better accuracy than just using the end-to-end learned model. 

\change{In all the experiments to follow (except for the ones analyzing the parameters for SMAID), we use the following settings. We use a VGG-16 model pre-trained on UCF101 dataset, to fine-tune the models for JHMDB, MPII Cooking activities, and the HMDB-51 datasets. As alluded to above, we use single RGB images for the RGB stream, a stack of ten consecutive optical flow images for the flow stream, and three-channel 21--45 frames summarized into SMAID images for the respective CNN stream. To train the SMAID CNN stream, we use the RGB stream of the above pre-trained model for initialization of the stream weights, which seemed to perform significantly better than learning from scratch. For fine-tuning, we used a fixed learning rate of $10^{-4}$ and a momentum of 0.9. We used the Caffe toolbox\footnote{\url{http://caffe.berkeleyvision.org/}} for our CNN implementations. We also applied the standard data augmentation techniques (such as mirroring) on the data inputs. For the RGB stream, the CNN iterations usually converged in about 20k iterations, the optical flow stream 40--60k iterations, and about 70k iterations for the SMAID stream. We also followed the same procedure for the ResNet-152 model by fine-tuning on a UCF101 pre-trained model.\footnote{\small{The VGG-16 and ResNet-152 pre-trained models are publicly available at \url{http://ftp.tugraz.at/pub/feichtenhofer/tsfusion/models/twostream_base/vgg16/} \\\url{http://ftp.tugraz.at/pub/feichtenhofer/tsfusion/models/twostream_base/resnet152/}.}}}

During testing, predictions from each of the three streams (output of FC8 layer in VGG-16 and the FC layer in ResNet-152), are normalized to be in $[0,1]$ after subtracting the minimum value, and are aggregated at the sequence level, kernelized (using a $\gamma=1$), and later vectorized after taking the matrix logarithm. For the MPII dataset, we used the provided training and validation sets. For JHMDB, we used 95\% of the training set to fine-tune CNNs, 5\% as validation. For the UCF101 dataset, we directly used  the pre-trained CNN models for the RGB and FLOW streams. For HMDB dataset, we trained our three streams by fine-tuning those used for UCF101. \change{ Note that while we use the pre-trained models from~\cite{feichtenhofer2016convolutional}, we do not use their fusion architecture in our evaluations. Instead, we use the setup in~\cite{simonyan2014two}, but using a VGG-16 or ResNet-152 model.}

 \begin{figure}
	\centering	
	\subfigure[]{\label{fig:smaid-a}\includegraphics[width=4cm,trim={0cm 6cm 0cm 5cm},clip]{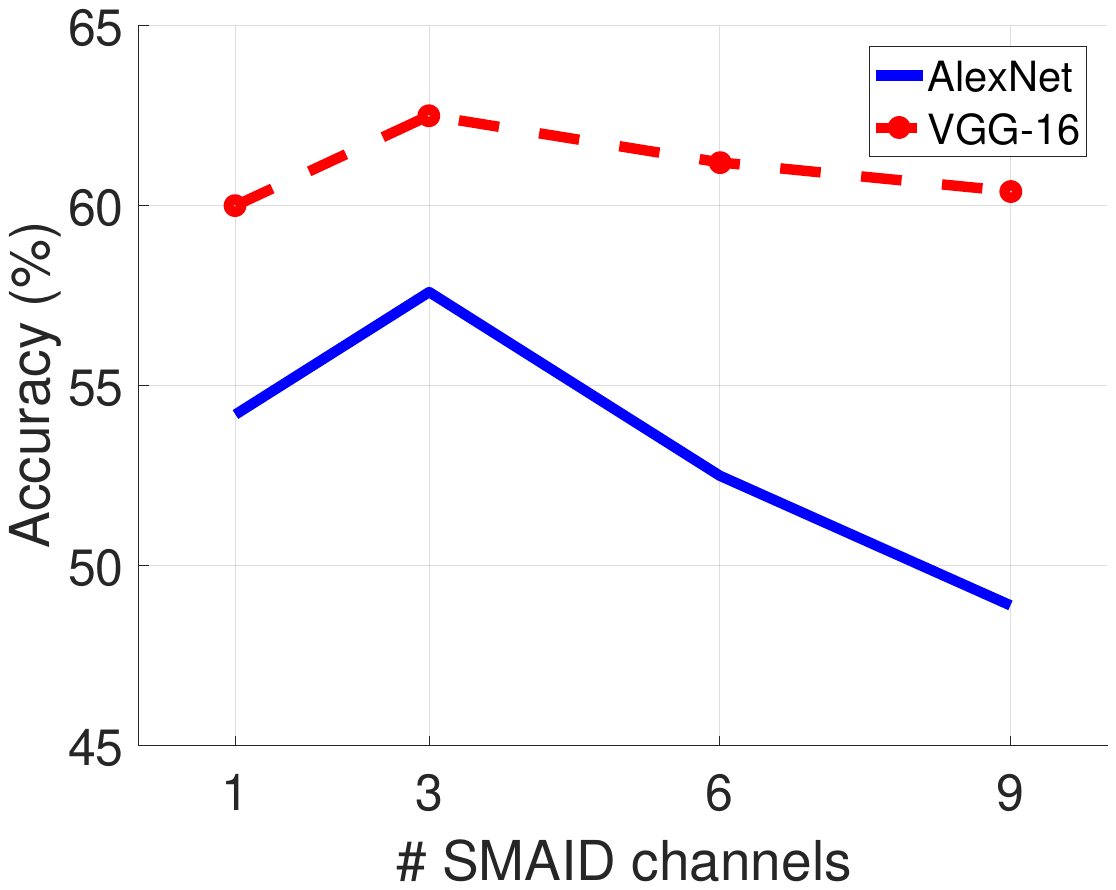}}					
	\subfigure[]{\label{fig:smaid-b}\includegraphics[width=4cm,trim={0cm 6cm 0cm 5cm},clip]{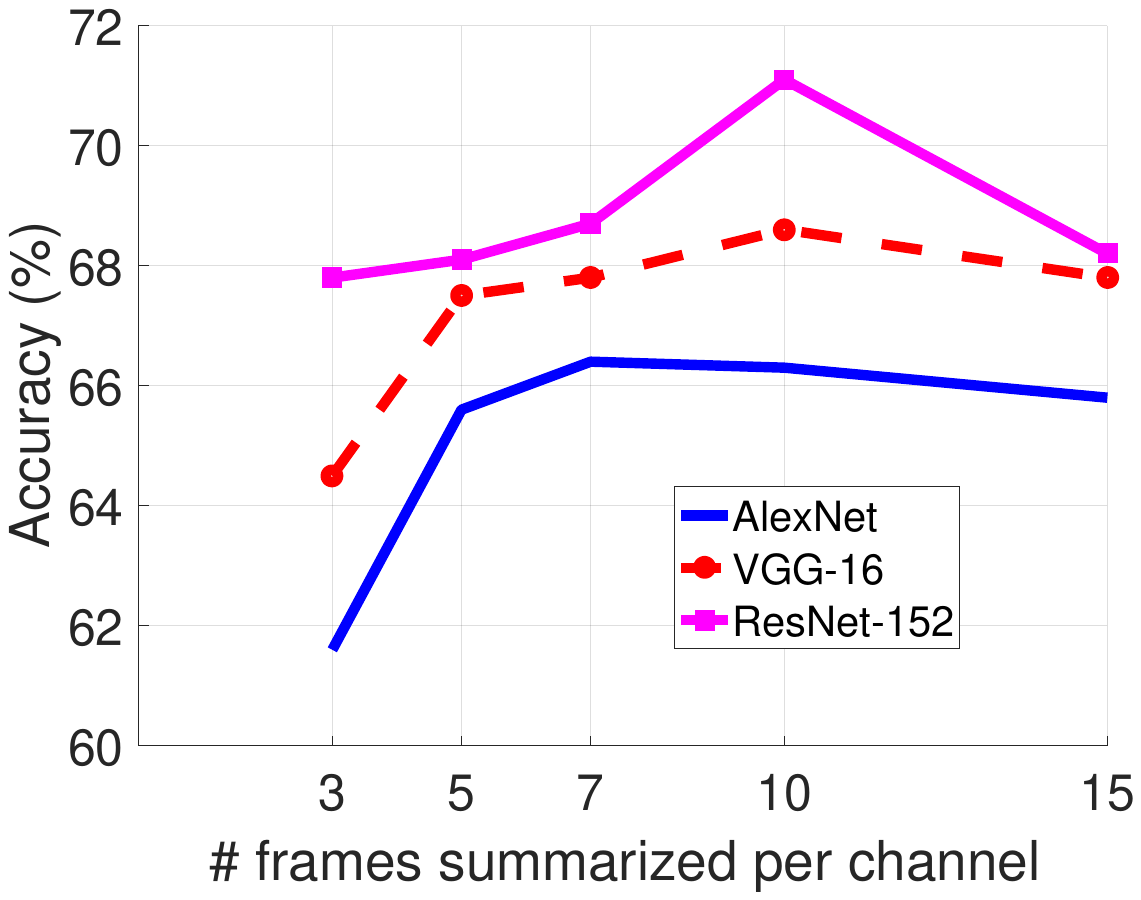}}
	\caption{\change{Evaluation of the effect of increasing number of SMAID channels (keeping number of frames per channel fixed at 7) and increasing number of frames per channel (while keeping the number of channels fixed at 3) on the UCF101 dataset split-1 using Alexnet, VGG-16, and ResNet-152 architectures.}\newchange{ Alexnet and VGG-16 was trained from scratch, while ResNet-152 was fine-tuned from an ImageNet pre-trained model.}}
	\label{fig:smaid_parameter_evaluation}
\end{figure}

\begin{figure*}[]
	\centering					
	\subfigure[MPII-VGG16]{\label{fig:bkcp-a}\includegraphics[width=5cm,trim={0cm 6cm 0cm 7cm},clip]{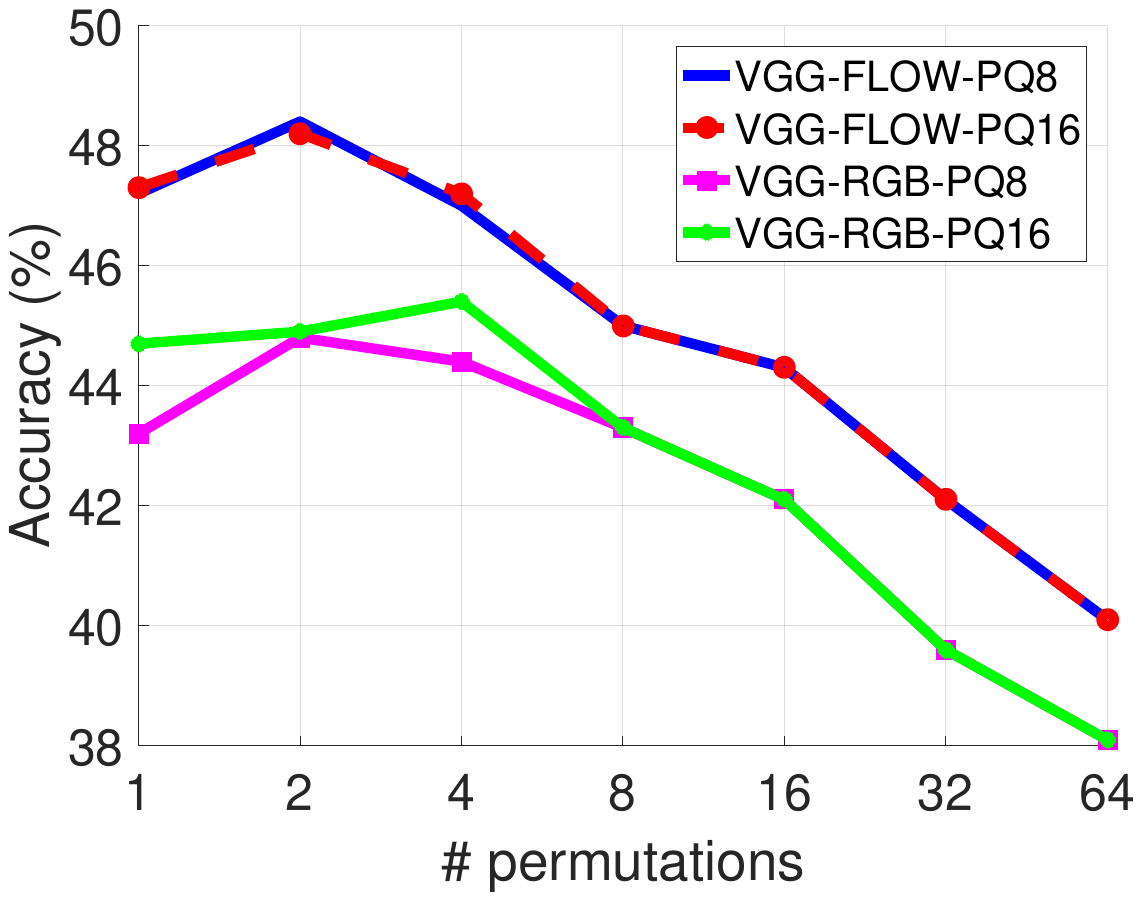}}
	\subfigure[UCF101-VGG16]{\label{fig:bkcp-b}\includegraphics[width=5cm,trim={0cm 6cm 0cm 7cm},clip]{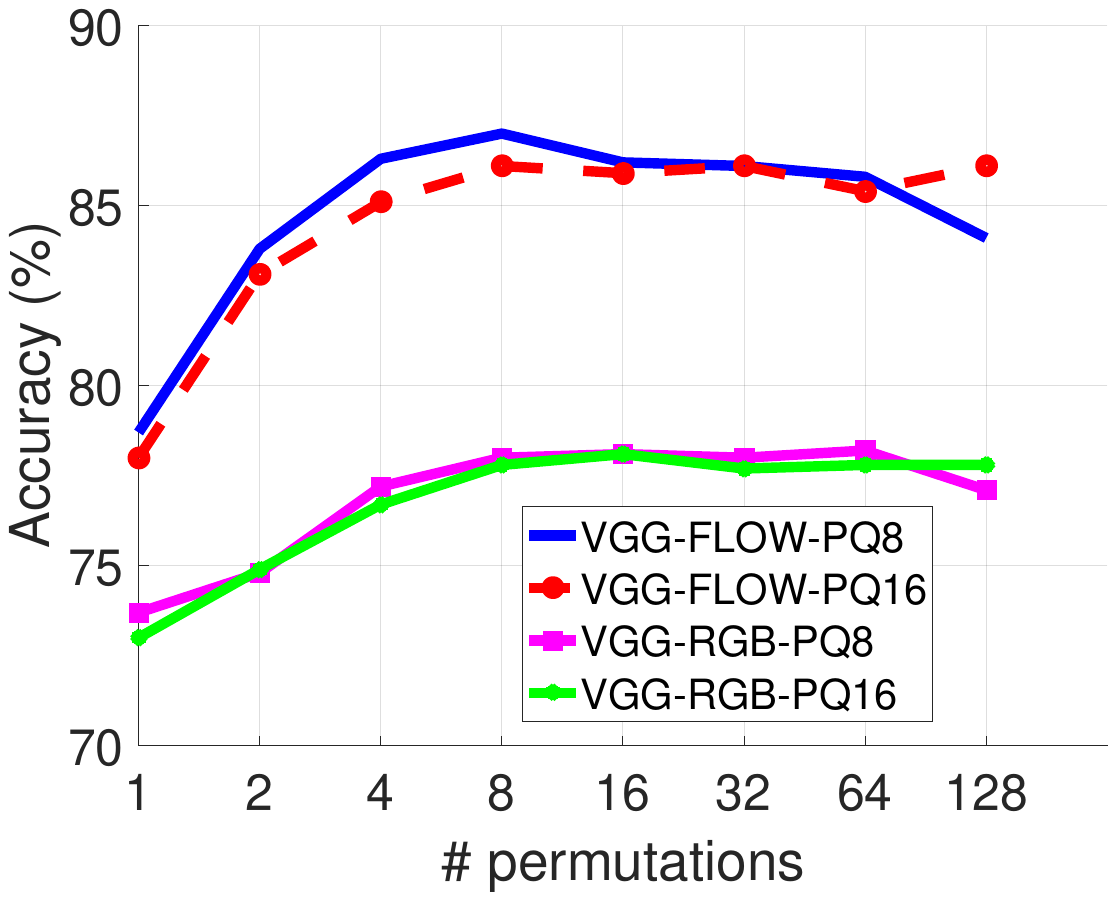}}	
    \subfigure[HMDB-ResNet-152]{\label{fig:bkcp-c}\includegraphics[width=5cm,trim={0cm 6cm 0cm 7cm},clip]{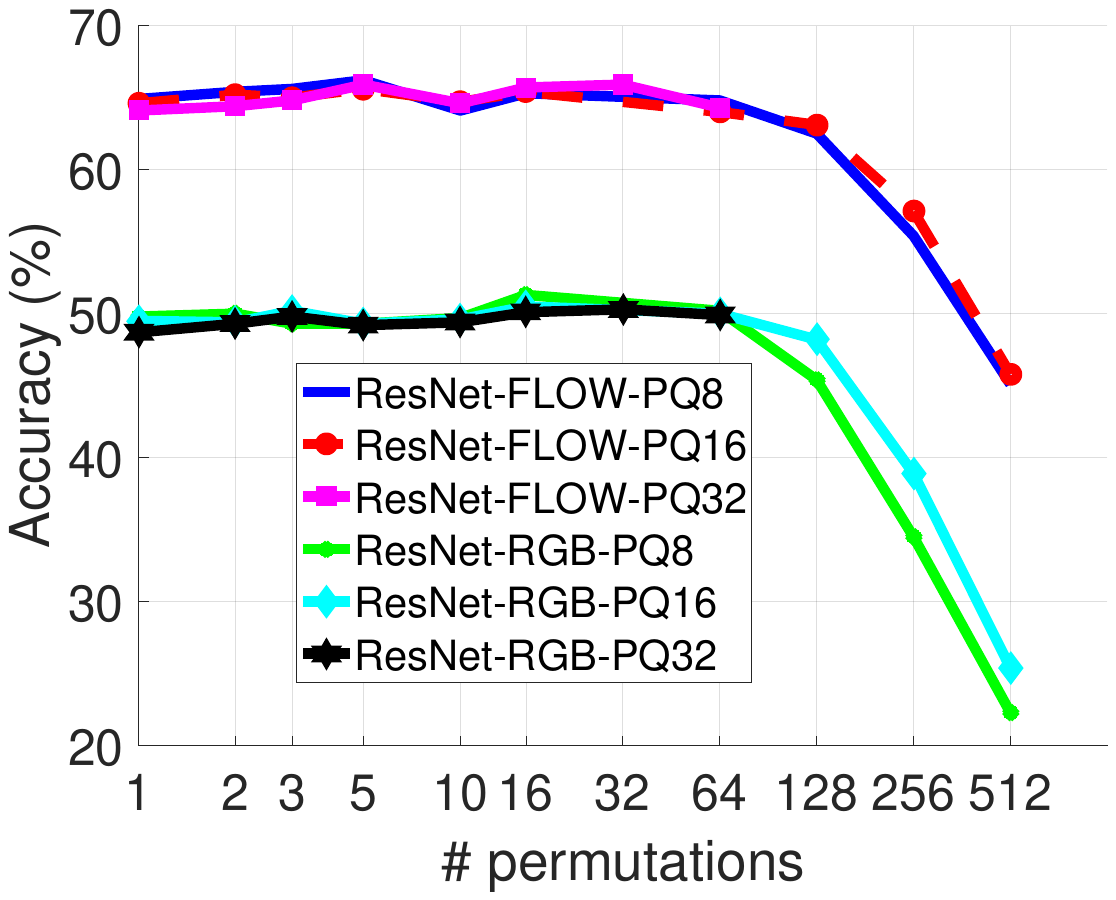}}    
	\caption{\change{Evaluation of the effect increasing the number of BKCP sub-vector dimensions and the size of the permutation set when constructing BKCP descriptors. Figure~\ref{fig:bkcp-a} shows the results on the MPII dataset using VGG16 fc6 features for sub-vector dimensionality 8 and 16 for optical flow and RGB streams. Figure~\ref{fig:bkcp-b} shows the same for the UCF101 dataset, and Figure~\ref{fig:bkcp-c} shows the results using ResNet-152 pool5 features on the HMDB dataset. The plots are generated after averaging the performance of 5 different random permutation sets for each dataset. The standard deviation of the performances is found to be less than 1\% on all the datasets. Best viewed in color.}
    }
	\label{fig:bckp_eval}
\end{figure*}

\subsection{SMAID Image Parameters}
As noted earlier, SMAID images summarize long-range actions into a compact image representation. There are two parameters for this representation: (i) number of frames that can be effectively summarized in a SMAID channel ($\zeta$ in~\eqref{eq:said}), and (ii) number of channels that can be stacked to capture the dynamics ($\beta$ in~\eqref{eq:ssaid}). Depending on the sequences, too many frame differences for (i) might result in a cluttered image that may not be useful for learning actions, while too less frames might lead to very sparse images. For (ii), while a 3-channel stack will render the SMAID as equivalent of an RGB image and thus RGB based CNN architectures could be used, higher-number of channels will require redesigning the network, and also leading to more CNN parameters. See Figure~\ref{fig:smaid_quals} for example frames from the UCF101 dataset for various number of frames encoded per channel in a 3-channel SMAID setup.

To understand the effect of these parameters, we progressively increased (i) and (ii) on a subset of the UCF101 split-1 training set containing videos that had limited camera motion, and evaluated on a small validation subset. 
{The plots use Alexnet, VGG16, and ResNet-152 models, the former two trained from scratch, while the ResNet-152 model is trained from an ImageNet pre-trained model \newchange{(as training from scratch takes too long due to the depth of the network).} In Figure~\ref{fig:smaid_parameter_evaluation}, we plot the classification accuracy. The plots reveal that higher number of frames per channel in SMAID leads to performance improvements, but with more than a certain number (for example, 7 for Alexnet, about 10 for VGG and ResNet), the performance drops, perhaps because of increasing clutter (see Figure~\ref{fig:smaid_quals}). On the other hand, with increasing number of SMAID channels (beyond three), the performance is seen to decrease for all the models, which is surprising. \change{We think this behavior is perhaps because of the typical network structure that we use, which is designed for RGB images, and is thus inadequate for a SMAID image with more than three channels. In the sequel, we use a 3-channel SMAID stack, with 15 frames per channel for the UCF101 dataset. We chose 15 frames instead of 10 frames as suggested from Figure~\ref{fig:smaid-b} because the difference between 7 or 10 frames per channel and 15 frames per channel in Figure~\ref{fig:smaid-b} is only about 1\%. Further, 15 frames per channel gives a longer 45 frames summarization of the sequence than say 30 frame-summarization using 10 frames per channel. For the ResNet-152 model, we use 10 frames per channel as the difference to 15 frames per channel is more than 5\%. As it is computationally expensive to cross-validate the best SMAID parameters for all our datasets, we repeated these parameter search experiments only for a few discrete settings and choose the best results in our subsequent experiments. We found that the same UCF101 parameters works well for the HMDB dataset. However, we found 7 frames per channel work best for MPII cooking activities and JHMDB datasets. With these configurations, each SMAID image captures subsequences of 45 frames in UCF101 and HMDB-51, and 21 frames in JHMDB and MPII datasets. }

We would also like to point out that SMAID with only one frame-difference per channel is equivalent to some of the recent proposals described in~\cite{Wang2016} and~\cite{sun2014discover}. However, as is clear from Figure~\ref{fig:smaid-b}, more frames per channel is significantly better. Further, looking back at Figure~\ref{fig:smaid-a}, a single channel SMAID is a grayscale image, similar to a motion history image~\cite{davis1997representation}. However, using more channels is clearly beneficial. These two plots substantiates that the design of SMAID is better than existing frame summarization techniques based on frame differencing. 
\change{
\begin{table*}[htbp]
	\centering
	\begin{tabular}{c|c|c|c|c|c|c|c|c}
		Experiment &  \small{MPII-mAP (\%)}  & \small{JHMDB-Avg.Acc.(\%)} & \multicolumn{2}{c|}{\small{HMDB-Avg. Acc.(\%)}} & \multicolumn{2}{c|}{\small{UCF101-Avg. Acc. (\%)}}\\        
        &  VGG16 & VGG16 & VGG16 & ResNet152 & VGG16 & ResNet152\\
		\hline        
		RGB   & 33.9  &  51.5 & 40.9 & 45.4  & 82.0 & 83.1\\
		FLOW    & 37.6&  54.8 & 47.5 & 59.5 &  85.1 &86.4\\
		SMAID  & 35.4  &  61.1 & 41.1 & 42.3 & 72.1 & 70.1\\
		RF     & 38.1  &  55.9 & 53.6 & 62.1 & 88.5 & 89.5\\
        RS     & 38.4 & 62.0 & 50.1  & 55.5 & 85.5 &  86.7\\
		RFS    & \textbf{39.5}  &  \textbf{62.6} & 54.4 & \textbf{63.5}  & 88.8 & \textbf{91.0}
	\end{tabular}
	\caption{\change{Sequence level comparison by directly averaging the CNN classifier outputs (not using correlation pooling) on MPII cooking dataset, JHMDB, HMDB and UCF101 (split-1). \change{The goal of this experiment is to show the advantages afforded by the standalone SMAID representation without any correlation pooling.} The results on MPII and JHMDB datasets use a VGG-16 architecture, while HMDB and UCF101 use both VGG-16 and ResNet-152 models.}}
	\label{tab:ssaid_mpii_jhmdb_frame_level_comparison}
\end{table*}
}

\subsection{Parameters for BKCP}
The block-diagonal approximation for KCP has two parameters, namely (i) the length of the sub-vectors ($p$ defined in~\eqref{eq:bkcp}) and the number of feature permutations to be tried to estimate the BKCP descriptor. For the former, as is clear, higher values of $p$ demands higher computations, while lower $p$ will ignore important correlations; in the limit $p=1$ is only the diagonal correlation matrix, which corresponds to average pooling. \change{In Figure~\ref{fig:bckp_eval}, we evaluate performance for various choices of these BKCP parameters on MPII cooking activities (Figure~\ref{fig:bkcp-a}), UCF101 (Figure~\ref{fig:bkcp-b}), and HMDB datasets (Figure~\ref{fig:bkcp-c}), using a VGG architecture, on the first two, and a ResNet-152 architecture on the third one. We mainly use $p=8,16$ as higher values lead to higher-dimensional descriptors (and thus are expensive, see Section~\ref{sec:complexity}), and also show inferior performance (Figure~\ref{fig:bkcp-c}). The latter observation is perhaps due to the fact that such higher dimensional sub-vectors result in mostly ill-conditioned blocks in TCP. This is because in most of our datasets, there are one average 50 to 100 frames per sequence. Given that we use the rectified features (after ReLU in the CNNs), they are mostly sparse. Both these factors result in TCP descriptors that are low-rank for higher $p$ and thus performance degrades. Thus, we find that using sub-vectors of length $p=16$ show good performance overall, and we use this configuration in our experiments to follow.}

\change{
From the plots in Figures~\ref{fig:bkcp-a},~\ref{fig:bkcp-b}, and~\ref{fig:bkcp-c}, we also find that a small number of permutations (in the range of 2--8) is sufficient to get a reasonable accuracy on all the datasets, and a higher number hurts. This suggests that the CNN features are perhaps strongly localized in their dimensions, as seen in the plots for a unit permutation set size. Further, we also suspect that averaging over too many randomized cross-dimensional correlations essentially marginalizes out any useful localized cues, thereby leading to poor accuracy. To validate this, we analyzed the average variance of the TCP descriptors for increasing number of permutations. We found that the variance steadily increases for more permutations. For example, it is on average 0.31 for ResNet-152 features for a single permutation, and goes beyond 1.5 when using 32 permutations. With such large variance, the data becomes mostly noise and thus any useful representational benefits are lost -- as is clear from the performance drop witnessed in Figure~\ref{fig:bkcp-c}. Thus, we use a permutation set of size 3 for VGG and 8 for ResNet152, in our subsequent experiments.
}

\comment{
As for the SMAID images, we used a subsequence length of seven frames, rescaled to range 0--255, with a three-channel stack which is equivalent to an RGB image. Such a representation has an effective temporal receptive field of 21 frames. We experimented with various lengths of the subsequences; we found that a few frames such as 3 may not capture any useful temporal cues, while too many frames lead to clutter in each MAID image. We repeated the above steps for the JHMDB dataset, except that we used a smaller MAID subsequence length of three frames, as these frames are already subsampled and contain significant motion. 
}

\section{Results}
In this section, we provide systematic evaluations of our various schemes on the four datasets. The notation RGB, FLOW, and SMAID denote the respective frame-level features. \change{We denote the combinations of RGB+FLOW as RF, RGB+SMAID as RS, and RGB+FLOW+SMAID as RFS, where the combinations are either averaged over their softmax CNN outputs for frame-level predictions, or their log-mapped features concatenated when using the correlation pooling schemes.}

\subsection{Evaluating the SMAID Representation} \change{First, we evaluate our SMAID representation at the frame-level against alternatives such as (i) using only a single stream image model RGB and (ii) using only optical flow stream FLOW. In Table~\ref{tab:ssaid_mpii_jhmdb_frame_level_comparison}, we provide these results on the four datasets. As is clear, SMAID is seen to improve performance on all the datasets, while its benefits are more on the MPII and JHMDB datasets (for example, the improvements from RF to RFS are about 2\% on MPII, and 8\% on JHMDB) as the camera motion is absent. While, the significance of SMAID is marginal on HMDB and UCF101 datasets  -- that have strong camera motions -- when using a VGG-16 architecture, we find that they show about 2\% improvement when using a powerful ResNet-152 model, which is encouraging. We also find that RS provides strong complementarity to the RGB stream (RGB to RS is 33.9 to 38.4 on MPII, 51.5 to 62.0, 45.4 to 55.5 on HMDB, and 83.1 to 86.7 on UCF101) showing about 6-10\% improvements. However, as is expected SMAID cannot replace the performance brought out by optical flow as is clear from the table.}


\comment{
\begin{table}[]
    \centering
    \begin{tabular}{c|c|c}
        Experiment & \small{MPII-mAP (\%)}  & \small{JHMDB-Avg.Acc. (\%)}\\
        \hline
         Im (image alone)  & 33.9  & 51.5\\
         Io (flow alone)   & 37.6  & 54.8\\
         Is (SMAID alone)  & 35.4  & 61.1\\
         Im + Io            & 38.1  & 55.9\\
         Im +Io + Is        & 39.5  & 62.6\\
    \end{tabular}    
    \caption{Sequence level comparison on MPII and JHMDB datasets. 
    	}
    \label{tab:ssaid_mpii_jhmdb_seq_level_comparison}
\end{table}
}

\begin{table}[htbp]
	\centering
	\begin{tabular}{c|c|c|c|c|c|c}
		Expt & \multicolumn{3}{c|}{\small{MPII}} & \multicolumn{3}{c}{\small{JHMDB}}\\
        \cline{2-7}
		&   \small{TCP} & \small{KCP} &  \small{BKCP}&  \small{TCP} & \small{KCP} & \small{BKCP}\\
		\hline
		RGB     & 49.7 & 52.7 & 55.2 &  44.8  & 51.8 & 48.8 \\
		FLOW    & 55.6 & 60.6 & 61.4 &  56.0  & 61.9 & 66.0  \\
		SMAID   & 51.3 & 55.7 & 59.6 &  47.2  & 59.7 & 55.6 \\
		RF      & 60.0 & 64.4 & 65.6 &  59.1  & 61.2 & 70.1 \\ 
        RS      & 57.2 & 61.9 & 64.9 &  49.1   & 60.1 & 63.4 \\
		RFS     & 62.1 & 66.1 & \textbf{68.0} &  62.1  & 72.4 & \textbf{73.6} \\ 
	\end{tabular}
	\caption{Comparison of temporal correlation pooling (TCP) against KCP and BKCP variants. These results use a VGG-16 model for the two datasets. MPII results use mAP scores and JHMDB reports the mean classification accuracy. All results are on the split-1.}
	\label{tab:mpii_jhmdb_cp_kcp_comparison}
\end{table}

\begin{table*}[htbp]
	\centering
	\begin{tabular}{c|c|c|c|c|c|c|c|c|c|c|c|c}
		\multirow{3}{*}{\small{Expt}} & \multicolumn{6}{c|}{\small{HMDB}} & \multicolumn{6}{c}{\small{UCF101}}\\   
        \cline{2-13}
		&   \multicolumn{2}{c|}{\small{TCP}} & \multicolumn{2}{c|}{\small{KCP}} &  \multicolumn{2}{c|}{\small{BKCP}}&  \multicolumn{2}{c|}{\small{TCP}} & \multicolumn{2}{c|}{\small{KCP}} & \multicolumn{2}{c}{\small{BKCP}}\\        
        \cline{2-13}
        & \small{VGG} & \small{ResNet} & \small{VGG} & \small{ResNet} & \small{VGG} & \small{ResNet} & \small{VGG} & \small{ResNet} & \small{VGG} & \small{ResNet} & \small{VGG} & \small{ResNet}\\
		\hline
		RGB     & 52.8 & 55.1& 56.7 & 59.9& 58.7 & 60.3 & 79.1 & 82.6 & 82.2 & 86.2 & 76.9 & 83.9\\
        FLOW    & 45.9 & 59.3& 53.3 & 65.2& 57.2 & 65.3 & 83.1 & 86.1 & 86.1 & 88.9 & 83.4 & 88.3\\   
        SMAID   & 49.4 & 24.9& 52.9 & 37.4& 52.1 & 43.7 & 74.2 & 72.2 & 71.7 & 75.3 & 70.7 & 71.2\\
        RF      & 57.1 & 63.6& 65.2 & 69.9& 68.1 & 69.6 & 86.2 & 91.1 & 87.8 & 94.0 & 87.5 & 92.9\\
        RS      & 55.2 & 56.7& 60.5 & 61.4& 63.0 & 61.3 & 82.2 & 85.1 & 85.7 & 86.9 & 81.2 & 85.0\\
        RFS     & 57.8 & 57.3& 66.7 & 68.4& 68.5 & \textbf{71.3} & 87.2 & 91.1 & 88.3 & \textbf{94.5} & 87.9 & 93.5\\			
	\end{tabular}
	\caption{\change{Comparison of classification accuracy between TCP, KCP, and BKCP. TCP and KCP uses the outputs of FC8 layers for VGG-16 and FC layer for ResNet-152. BKCP uses FC6 layer features of VGG-16 and pool5 features of ResNet-152. All results are on the split-1.}}
	\label{tab:hmdb_and_ucf101_cp_kcp_comparison}
\end{table*}

\subsection{Correlation Pooling} Next, we evaluate our correlation pooling (TCP) scheme and its kernelized variants (KCP and BKCP) on CNN features (\change{FC8 and FC layer for VGG16 and ResNet-152 respectively for both TCP and KCP, and FC6 and pool5 of VGG16 and ResNet-152 respectively for BKCP}) from the three input modalities. The results are shown in Tables~\ref{tab:mpii_jhmdb_cp_kcp_comparison} and~\ref{tab:hmdb_and_ucf101_cp_kcp_comparison}. Comparing these results to those in Table~\ref{tab:ssaid_mpii_jhmdb_frame_level_comparison}, show that KCP improves sequence level  performance substantially on all datasets; from 39.5\% to 66.1\% for RFS on MPII, from 62.6\% to 72.4\% on JHMDB, from 54.4\% on HMDB-51 63.5 to 66.5\% and 91.0\% to 94.5\% on UCF101. Tables~\ref{tab:mpii_jhmdb_cp_kcp_comparison} and~\ref{tab:hmdb_and_ucf101_cp_kcp_comparison} also show that kernelizing the temporal correlations (TCP versus KCP or BKCP) is always useful; demonstrating a consistent 5-10\% improvement from its non-kernelized variant. We also find that BKCP performs better than KCP overall. \change{This is unsurprising given that BKCP has more dimensionality, and also captures features in the CNN pipeline more closer to the input images than KCP that directly operates on the CNN classifier outputs -- as a result, class confusions are more prominent in BKCP and thereby better correlation descriptors. However, for UCF101 dataset, we find that the effects are reversed almost consistently. This we suspect is due to the larger training size of this dataset -- as a result the final CNN features are already very discriminative for the actions. This intuition is consistent with the results in Table~\ref{tab:ssaid_mpii_jhmdb_frame_level_comparison} for UCF101, where the final frame-level average pooling accuracy is already high. However, we still find that KCP and BKCP is beneficial and improves the average pooling performance by about 2-3\%.}



\subsection{BKCP versus Low-Rank Decomposition}
\change{Recall that BKCP is introduced as KCP descriptors turned out to be too expensive for high-dimensional CNN features. However, an alternative would be to use a low-rank decomposition, such as PCA, on these features and then apply KCP on the low-dimensional features obtained after projection onto the principal components. We explore this alternative in Table~\ref{tab:pca_bkcp} on the MPII and HMDB datasets. For learning the basis, we randomly sample 1000 sequences from the respective training sets and their associated CNN features, followed by applying an SVD to find the basis. We tried various number of basis (based on the performance on a validation set) and selected 256 basis that seemed to give the best performance in terms of feature dimensionality, computational expenditure, and accuracy. As is clear from the Table~\ref{tab:pca_bkcp}, using PCA does provide useful lower dimensional KCP representations, however, BKCP still outperforms it. We think this is because as the basis are learned generically over a large portion of the dataset, the sequence level features when projected onto such a basis, may lose information that are perhaps subtle (and thus not captured by any principal component) and important. We see a consistent drop in performance on both RGB and FLOW streams for both the datasets. 
}

\subsection{KCP versus Fisher Vectors}
\change{In this section, we compare KCP with Fisher vector encodings, which are well-known and successful second-order representations used in a variety of vision applications, including action recognition~\cite{newell2016stacked}. In this experiment, we apply Fisher vectors on the output of the last CNN layer (as is used in for generating KCP descriptors). A first step to generate Fisher vectors is to train a Gaussian Mixture model. To this end, similar to our approach in the last section, we sampled 1000 sequences randomly from the respective training set, and used 256 Gaussians in the mixture. Once the mixture model is trained, we used the VLFeat software~\footnote{\url{http://www.vlfeat.org}} to generate Fisher vector encodings for every sequence, which is then classified using a linear classifier. Our results and comparisons to exactly similar features represented using KCP descriptors is provided in Table~\ref{tab:fv_kcp}. Again, as observed in the previous section, we see a significant drop in performance when using Fisher vectors against KCP on both HMDB and MPII datasets and for both FLOW and RGB modalities.
}
\begin{table}[htbp]
	\centering
	\begin{tabular}{c|c|c}
		Experiment &  \small{HMDB-Avg. Acc.(\%)} & \small{MPII mAP (\%)}\\
		\hline        
        R-BKCP &  \textbf{52.1} & \textbf{55.2}\\
        F-BKCP& \textbf{66.0} & \textbf{61.4}\\
        \hline
        R-PCA& 46.1 &  40.3\\
        F-PCA& 64.8 & 48.8 \\        
	\end{tabular}
	\caption{Comparison of BCKP to low-rank decompositions (PCA). For the latter, we use 256 basis, and applied KCP on the resulting projected features. HMDB uses ResNet-152 pool5 features (2048D), while MPII uses VGG16 features (4096D).}
	\label{tab:pca_bkcp}
\end{table}

\begin{table}[htbp]
	\centering
	\begin{tabular}{c|c|c}
		Experiment &  \small{HMDB-Avg. Acc.(\%)} & \small{MPII mAP (\%)}\\
		\hline        
		R-KCP & \textbf{49.9} &  \textbf{52.7}\\
        F-KCP & \textbf{65.2} & \textbf{ 60.6}\\
        \hline
        R-Fisher Vec & 35.9 & 30.4 \\
        F-Fisher Vec & 48.7 & 38.1 \\      		
	\end{tabular}
	\caption{Comparison of KCP to Fisher vector encoding. We used 256 Gaussians in the Fisher vector Gaussian mixture model. HMDB and MPII datasets use the outputs of the last fully-connected layers of a ResNet-152 and VGG-16 models respectively.}
	\label{tab:fv_kcp}
\end{table}

\subsection{KCP Classification Kernel}
\label{sec:kernel_comparison}
As reviewed in Section~\ref{sec:kernels}, there are popularly two SVM kernels on SPD matrices, the Stein kernel and the Log-Euclidean kernel. In Table~\ref{tab:kernel_comparison}, we show results comparing these two kernels on the MPII Cooking Activities and the JHMDB datasets. As is clear, either kernel performs differently and generate improvements, suggesting that it is better to cross-validate each of the kernels on the respective datasets to choose the right one. Given that the improvements produced by the log-euclidean kernel on the JHMDB dataset is significantly higher than the improvements by the Stein kernel on the MPII dataset and further noting the computational advantages as described in Section~\ref{sec:log-euc}, we decided to use the log-euclidean kernel in the sequel.

\begin{table}[]
	\centering
	\begin{tabular}{c|c|c|c|c}
		Experiment & \small{MPII}& \small{JHMDB} \\
		& \small{KCP mAP(\%)} & \small{KCP Avg.Acc.(\%)}\\
		\hline                   
		LE Kernel        & 66.1 &  72.7 \\
		Stein kernel  & 68.5  & 62.5 \\
	\end{tabular}
	\caption{Comparison of performance when using different kernels in SVM for classifying the kernelized correlation matrices over the three input modalities.}
	\label{tab:kernel_comparison}
\end{table}

\begin{table}
\centering
\small
\begin{tabular}{l|c|c|c}
Class Name & \# seq & ~\cite{zhou2015interaction} & Ours\\ 
\hline
Change temperature & 27 & 59.26 & \textbf{96.30}\\ 
Cut apart & 97 & 50.52 & \textbf{62.89}\\ 
Cut dice & 40 & 12.50 & \textbf{22.50}\\ 
Cut in & 12 & \textbf{25.00} & 0.00\\ 
Cut off ends & 27 & \textbf{48.15} & 3.70\\ 
Cut out inside & 37 & 62.16 & \textbf{75.68}\\ 
Cut slices & 91 & 40.66 & \textbf{81.32}\\ 
Cut stripes & 12 & \textbf{25.00} & 16.67\\ 
Dry & 26 & 92.31 & \textbf{100.00}\\ 
Fill water from tap & 3 & \textbf{100.00} & 66.67\\ 
Grate & 19 & 63.16 & \textbf{78.95}\\ 
Lid: put on & 6 & \textbf{50.00} & 0.00\\ 
Lid: remove & 8 & \textbf{87.50} & 0.00\\ 
Mix & 5 & \textbf{60.00} & 0.00\\ 
Move from X to Y & 70 & 72.86 & \textbf{75.71}\\ 
Open egg & 5 & \textbf{80.00} & 40.00\\ 
Open tin & 7 & 71.43 & \textbf{71.43}\\ 
Open/close cupboard & 18 & \textbf{88.89} & 66.67\\ 
Open/close drawer & 58 & 48.28 & \textbf{68.97}\\ 
Open/close fridge & 8 & \textbf{87.50} & 50.00\\ 
Open/close oven & 1 & \textbf{100.00} & 0.00\\ 
Package X & 6 & \textbf{83.33} & 16.67\\ 
Peel & 64 & 76.56 & \textbf{79.69}\\ 
Plug in/out & 6 & \textbf{100.00} & 33.33\\ 
Pour & 55 & \textbf{83.64} & 72.73\\ 
Pull out & 4 & \textbf{100.00} & 25.00\\ 
Puree & 12 & 75.00 & \textbf{83.33}\\ 
Put in bowl & 127 & 40.16 & \textbf{88.98}\\ 
Put in pan/pot & 28 & 32.14 & \textbf{75.00}\\ 
Put on bread/dough & 149 & 55.70 & \textbf{93.29}\\ 
Put on cutting-board & 57 & \textbf{63.16} & 45.61\\ 
Put on plate & 55 & 30.91 & \textbf{70.91}\\ 
Read & 8 & 50.00 & \textbf{50.00}\\ 
Remove from package & 15 & \textbf{60.00} & 53.33\\ 
Rip open & 6 & \textbf{66.67} & 0.00\\ 
Scratch off & 12 & \textbf{58.33} & 0.00\\ 
Screw close & 44 & \textbf{75.00} & 52.27\\ 
Screw open & 45 & \textbf{68.89} & 53.33\\ 
Shake & 72 & 73.61 & \textbf{83.33}\\ 
Smell & 16 & 12.50 & \textbf{56.25}\\ 
Spice & 20 & \textbf{80.00} & 55.00\\ 
Spread & 12 & \textbf{50.00} & 25.00\\ 
Squeeze & 18 & 66.67 & \textbf{83.33}\\ 
Stamp & 8 & 62.50 & \textbf{75.00}\\ 
Stir & 38 & 57.89 & \textbf{86.84}\\ 
Strew & 40 & 17.50 & \textbf{72.50}\\ 
Take \& put in cupboard & 10 & \textbf{80.00} & 30.00\\ 
Take \& put in drawer & 8 & \textbf{62.50} & 12.50\\ 
Take \& put in fridge & 9 & \textbf{100.00} & 66.67\\ 
Take \& put in oven & 3 & 100.00 & \textbf{100.00}\\ 
Take \& put in spice holder & 13 & 61.54 & \textbf{61.54}\\ 
Take ingredient apart & 39 & \textbf{48.72} & 43.59\\ 
Take out from cupboard & 57 & \textbf{94.74} & 92.98\\ 
Take out from drawer & 130 & 85.38 & \textbf{94.62}\\ 
Take out from fridge & 34 & 94.12 & \textbf{97.06}\\ 
Take out from oven & 3 & \textbf{100.00} & 0.00\\ 
Take out from spice holder & 17 & \textbf{82.35} & 70.59\\ 
Taste & 12 & \textbf{75.00} & 16.67\\ 
Throw in garbage & 39 & 64.10 & \textbf{84.62}\\ 
Unroll dough & 3 & 100.00 & \textbf{100.00}\\ 
Wash hands & 45 & \textbf{55.56} & 37.78\\ 
Wash objects & 91 & \textbf{96.70} & 86.81\\ 
Whisk & 9 & 77.78 & \textbf{88.89}\\ 
Wipe clean & 10 & \textbf{80.00} & 20.00\\ 
\hline
Mean & & 62.7 & \textbf{70.0}
\end{tabular}
\caption{A comparison of classification accuracy per class (not mAP) on the MPII dataset against the method in~\cite{zhou2015interaction}. Our method uses KCP.}
\label{tab:mpii_per_class_accuracy}
\end{table}
\comment{
\subsection{Alternative CNN Architectures}
In this subsection, we compare the performance reported in Table~\ref{tab:mpii_jhmdb_cp_kcp_comparison} when using an Alexnet model instead of a VGG-16 network. While, it is obvious that the performance will be inferior, our goal is to verify if the correlation pooling schemes demonstrate a similar trend. In Table~\ref{tab:alexnet_comparisons}, we show the results of this experiment. As is clear, using VGG-16 shows better results against Alexnet model in all the input modalities, and we also find that KCP is always better than TCP, thus confirming that the trend that we see is independent of the network architecture.

\begin{table}[]
	\centering
	\begin{tabular}{c|c|c}
		Experiment & \small{MPII}& \small{MPII} \\
		& \small{TCP mAP(\%)} & \small{KCP mAP (\%)}\\
		\hline                   
		RGB   & 44.7  &  48.0 \\
		FLOW    & 32.2 &  37.0 \\
		SMAID  & 50.1  &  53.3\\
		RF     & 46.4  &  50.9\\
		RFS    & \textbf{53.1}  &  \textbf{59.8} 
	\end{tabular}
	\caption{Improvements between TCP and KCP when using an Alexnet CNN model instead of a VGG-16 network (compare to Table~\ref{tab:mpii_jhmdb_cp_kcp_comparison} on the MPII Cooking activities dataset.}
	\label{tab:alexnet_comparisons}
\end{table}
}
\begin{table}[]
	\centering
	\begin{tabular}{c|c}
		Algorithm &  mAP(\%) \\
		\hline
		Holistic + Pose~\cite{rohrbach2012database} & 57.9 \\
		Video Darwin~\cite{fernando2015modeling}    & 72.0 \\
		Interaction Part Mining~\cite{zhou2015interaction} & 72.4 \\
		P-CNN~\cite{cheron2015p} & 62.3 \\
		P-CNN + IDT-FV~\cite{cheron2015p} & 71.4 \\
		Semantic Features~\cite{zhou2014pipelining} & 70.5 \\
		Hierarchical Mid-Level Actions~\cite{lan2015action} & 66.8\\
        Higher-order Pooling~\cite{hok} & 73.1\\       
		\hline
		KCP & 66.1\\
		BKCP  & 68.0\\ 
	    BKCP + KCP   & 68.6 \\
		KCP + Trajectories & 73.5\\
		BKCP + Trajectories & 72.4 \\
		BKCP + KCP + Trajectories & \textbf{74.7}\\
	\end{tabular}
	\caption{MPII Cooking Activities (7-splits)}
	\label{tab:mpii_soa}
\end{table}
\begin{table}
	\centering
	\begin{tabular}{c|c}
		Algorithm &  Avg. Acc. (\%) \\
		\hline
		P-CNN~\cite{cheron2015p} & 61.1 \\
		P-CNN + IDT-FV~\cite{cheron2015p} & 72.2 \\
		Action Tubes~\cite{gkioxari2015finding} & 62.5\\
		Stacked Fisher Vectors~\cite{peng2014action} & 69.03\\
		IDT + FV~\cite{wang2013action} & 62.8 \\
        Higher-order Pooling~\cite{hok} & 73.3\\     
		\hline
		KCP & 72.7\\
		BKCP & 72.4\\
		BKCP + KCP & 73.7 \\
		KCP + IDT-FV & 74.1\\
		BKCP + KCP + IDT-FV  & \textbf{77.3}\\
	\end{tabular}
	\caption{JHMBD Dataset (3-splits)}
	\label{tab:jhmdb_soa}
\end{table}
\begin{table}[]
	\centering
	\begin{tabular}{c|cc|cc|}
		Algorithm &  \multicolumn{2}{c|}{\small{HMDB-51(\%)}}  & \multicolumn{2}{c|}{\small{UCF101(\%)}}  \\                  
		\hline
		Two-stream \cite{simonyan2014two}                       & \mc{59.4} & \mc{88.0} \\
        Two-stream Fusion \cite{feichtenhofer2016convolutional} & \mc{69.2} & \mc{93.5}  \\
        TSN\cite{Wang2016}                						& \mc{69.4} & \mc{94.2} \\
        I3D (Kinetics)~\cite{carreira2017quo} & \mc{80.2} & \mc{97.9}\\
    	ActionVLAD+IDT~\cite{girdhar2017actionvlad}                        & \mc{69.8} & \mc{93.6}\\
		I3D+SVMP~\cite{wang2018video} & \mc{81.3} & \mc{--}\\
        Kernel Rank Pool~\cite{cherian2018non} &\mc{74.2} &\mc{--} \\
        IDT+FV \cite{wang2013action}                            & \mc{57.2} & \mc{85.9}\\
        IDT+HFV \cite{peng2016bag}                              & \mc{61.1} & \mc{87.9} \\
        TDD+IDT \cite{wang2015action}                           & \mc{65.9} & \mc{91.5} \\
        DT+MVSV \cite{cai2014multi}                             & \mc{55.9} & \mc{83.5} \\
        Dynamic Image \cite{bilen2016dynamic}          			& \mc{65.2} & \mc{89.1} \\
        ST-ResNet \cite{feichtenhofer2016spatiotemporal}        & \mc{70.3} & \mc{94.6} \\	
        ST-Multiplier\cite{feichtenhofer2017spatiotemporal}        & \mc{72.2} &  \mc{94.9}     \\
		\hline
         & \multicolumn{1}{c|}{\small{VGG}} & \multicolumn{1}{c|}{\small{ResNet}} &\multicolumn{1}{c|}{\small{VGG}} & \multicolumn{1}{c|}{\small{ResNet}}\\
         \cline{2-5}
		KCP                 & 65.8 & 68.7 & 89.1 & 91.0 \\
		BKCP                & 68.5 & 70.0 & 88.6 & 91.1\\
		KCP + BKCP          & 67.8 & 71.3 & 89.4 & 93.7\\
		KCP + IDT-FV        & 67.2 & 69.2 & 92.0 & 94.5\\
		BKCP + IDT-FV       & 69.6 & 72.0 & 89.3 & 93.1\\
		BKCP + KCP + IDT-FV & 70.5 & \textbf{72.5} & 92.4 & \textbf{95.4}\\
	\end{tabular}
	\caption{Average classification accuracy (\%) over 3-splits on the HMBD-51 and UCF-101 Datasets.}
	\label{tab:hmdb_soa}  
    \vspace*{-0.5cm}
\end{table}
\begin{table}[!h]
	\centering
	\begin{tabular}{c|c|c}
		Action & RF  & RFS  \\ 
		& mAP (\%)  & mAP (\%) \\
		\hline
		Change Temperature & 32.1 & 50.7 \\
		Dry   & 46.6 & 53.6 \\
		spice & 29.4 & 34.9 \\
		put on cupboard & 24.1 & 18.3 \\ 
	\end{tabular}    
	\caption{Analysis of per-class recognition accuracy on the MPII dataset with avg. pooling when using RGB + FLOW against RGB+FLOW+SMAID.
	}
	\label{tab:comparison_SMAID_table}
\end{table}
\begin{figure}[ht]
	\centering
	\includegraphics[width=8cm]{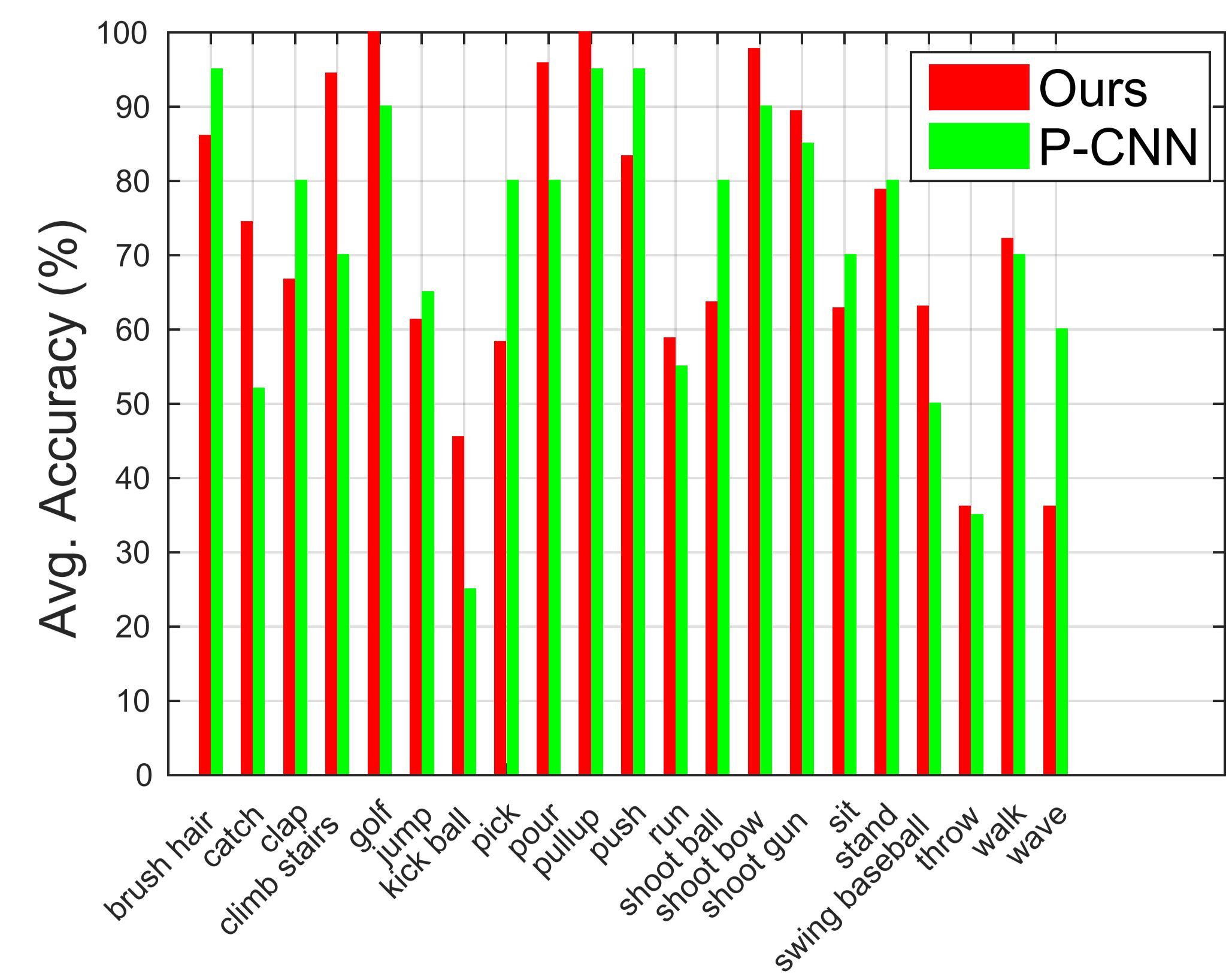}
	\caption{KCP per-class classification accuracy of sequences in the JHMDB dataset in comparison to the results in~\cite{cheron2015p}.}
	\label{fig:jhmdb_per_class_accuracy}
\end{figure}
\begin{figure*}[h]
    \centering
    \includegraphics[width=16cm,trim={0 8cm 0 1cm},clip]{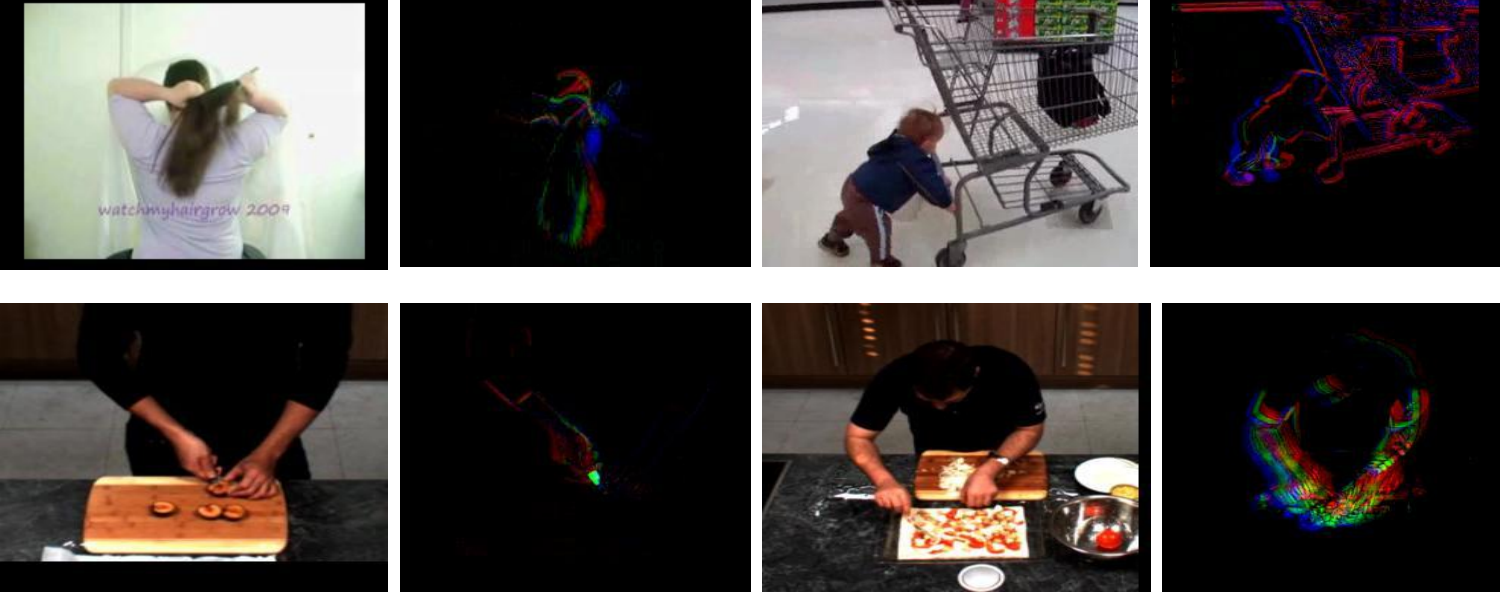}
    \caption{Qualitative SMAID and the associated appearance images from the JHMDB dataset (top) and the MPII Cooking activities dataset (bottom).}
    \label{fig:qual_results}    
\end{figure*}
\begin{figure*}[h]
\centering
\includegraphics[width=2cm]{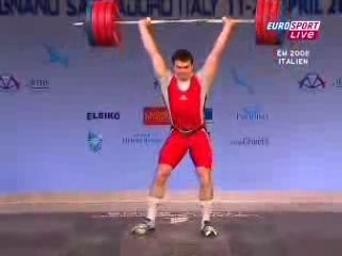}
\includegraphics[width=2cm]{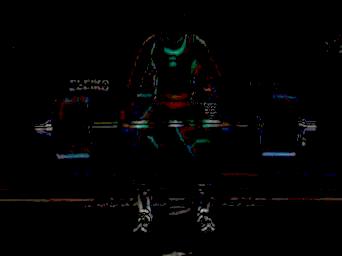}
\includegraphics[width=2cm]{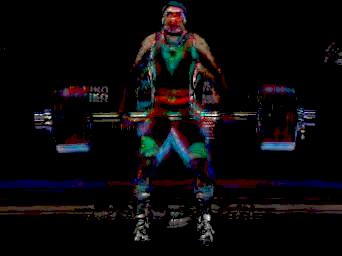}
\includegraphics[width=2cm]{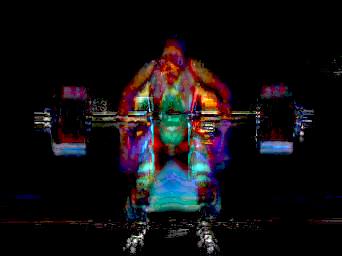}
\includegraphics[width=2cm]{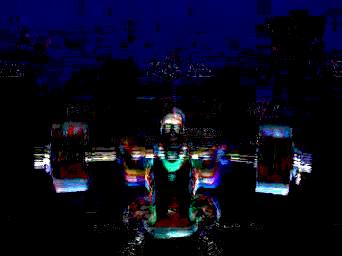}
\includegraphics[width=2cm]{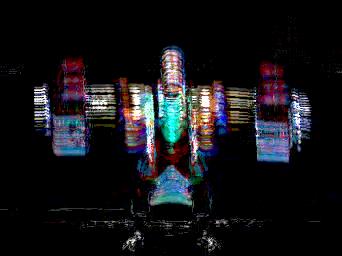}
\includegraphics[width=2cm]{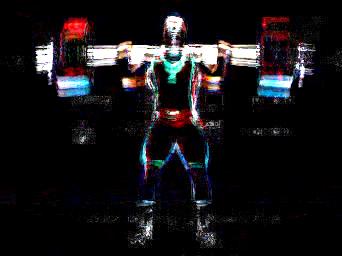}
\includegraphics[width=2cm]{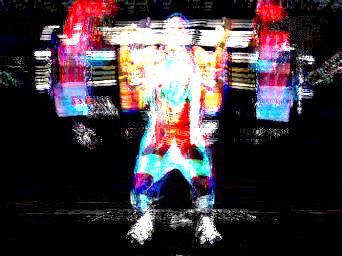}\\
\includegraphics[width=2cm]{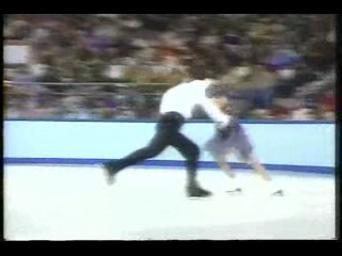}
\includegraphics[width=2cm]{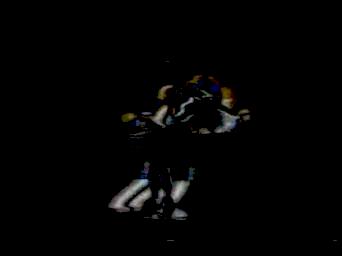}
\includegraphics[width=2cm]{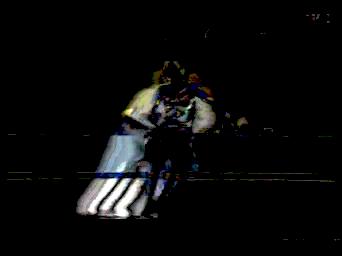}
\includegraphics[width=2cm]{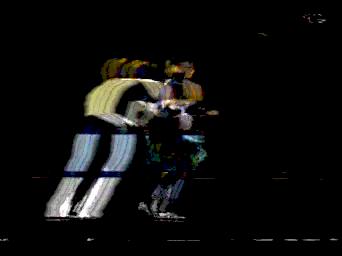}
\includegraphics[width=2cm]{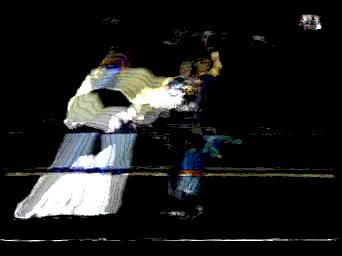}
\includegraphics[width=2cm]{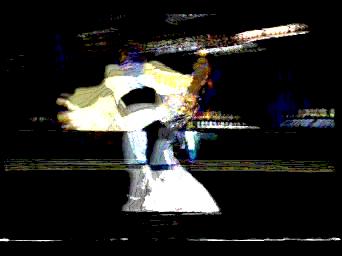}
\includegraphics[width=2cm]{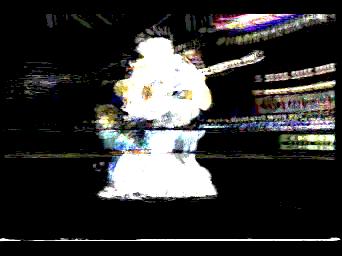}
\includegraphics[width=2cm]{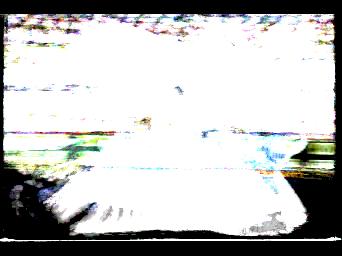}\\
\subfigure[RGB frame]{\includegraphics[width=2cm]{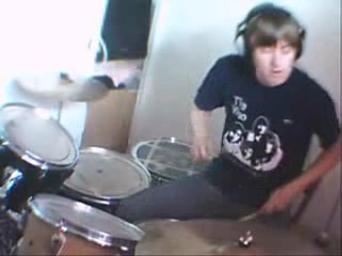}}
\subfigure[$\zeta$=2]{\includegraphics[width=2cm]{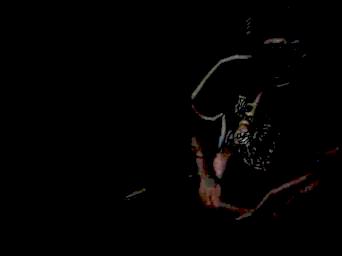}}
\subfigure[$\zeta$=3]{\includegraphics[width=2cm]{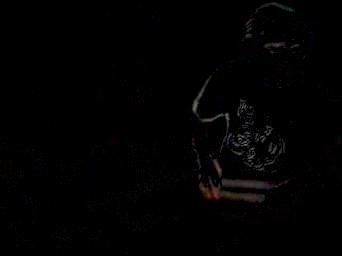}}
\subfigure[$\zeta$=5]{\includegraphics[width=2cm]{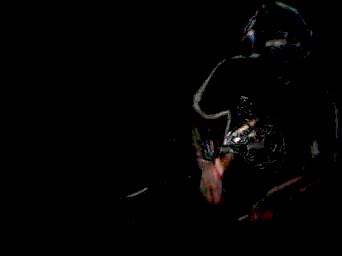}}
\subfigure[$\zeta$=10]{\includegraphics[width=2cm]{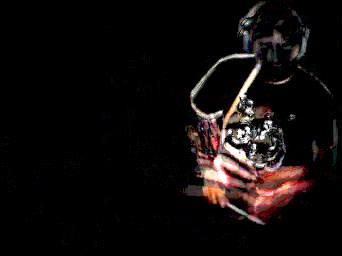}}
\subfigure[$\zeta$=15]{\includegraphics[width=2cm]{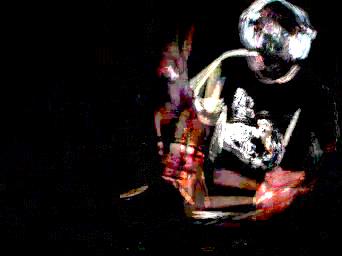}}
\subfigure[$\zeta$=30]{\includegraphics[width=2cm]{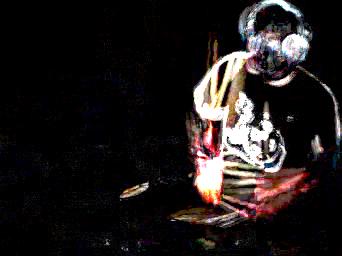}}
\subfigure[$\zeta$=60]{\includegraphics[width=2cm]{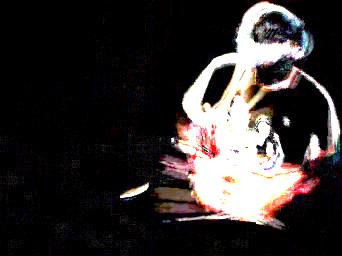}}
\caption{Comparison of 3-channel SMAID images with varying number of frames summarized per channel ($\zeta$) for two sequences from UCF101 dataset. As is clear, higher $\zeta$ leads to cluttered image, while smaller $\zeta$ fails to capture sufficient motion. Also, note that for each SMAID image, the temporal order is mapped to colors Red $<$ Green $<$ Blue.}
\label{fig:smaid_quals}
\end{figure*}

\subsection{Comparisons to the State of the Art} In Tables~\ref{tab:mpii_soa},~\ref{tab:jhmdb_soa}, and~\ref{tab:hmdb_soa}, we present comparisons of our full framework (RGB + FLOW + SMAID) against state-of-the-art approaches on the four datasets, averaging the performance on all splits. On the MPII cooking activities dataset, our kernelized correlation pooling scheme shows an overall mAP of 68\% (Table~\ref{tab:mpii_jhmdb_cp_kcp_comparison}).  This is better than the results in recent CNN based approaches such as~\cite{cheron2015p} (62.3\%) and better than non-CNN based, yet state of the art schemes such as~\cite{lan2015action} (66.8\%). Further, we see that incorporating trajectory features into our framework substantially improves our accuracy further to 74.7\% (Table~\ref{tab:mpii_soa}) outperforming all other approaches. On the JHMDB dataset, our correlation pooling scheme provides an average accuracy of 62\%, while the kernelization scheme improves this to 72.7\%. In comparison to the CNN based results in~\cite{cheron2015p}, our results are about 10\% better. Further, incorporating BKCP and trajectory features increases our performance to 77.3\%, which is better than the next best method by about 5.1\%. \change{These comparisons clearly demonstrate the effectiveness of our methodology against prior works. On HMDB dataset, our combination of KCP, BKCP, with dense trajectory features demonstrate state of the art performance, better by about 1.3\%, on a similar capacity VGG-16 model~\cite{feichtenhofer2016convolutional}, and providing about 2.2\% improvement over the respective ResNet model (70.3\% to 72.5\%) when combined with Fisher vector encoded trajectory features. Similar results are seen on UCF101 dataset, with our scheme outperforming a recent state of the art~\cite{feichtenhofer2017spatiotemporal} by 0.5\%}. We also provide comparison to the recent I3D model~\cite{carreira2017quo}, however note that this model was pre-trained using the larger Kinetics-400 dataset~\cite{zisserman2017kinetics}, and thus the results are not strictly comparable to those on other methods or ours (that do not use external dataset). From the tables, it is clear that our scheme is independent of the CNN architecture and is consistent in the improvements that it produces in comparison to first-order pooling schemes (Table~\ref{tab:ssaid_mpii_jhmdb_frame_level_comparison} and Table~\ref{tab:mpii_jhmdb_cp_kcp_comparison}). 

\subsection{Analysis of Results}
In this section, we provide more analysis of our results, summarizing when second-order methods improved the performance in the datasets that we use. In Figure~\ref{fig:jhmdb_per_class_accuracy} and Table~\ref{tab:mpii_per_class_accuracy}, we provide the accuracy of each class when using KCP as against those from state of the art methods on the JHMDB and the MPII Cooking activities datasets respectively. On the MPII dataset, we outperform~\citet{zhou2015interaction} on 28 sequences (out of 64), and in most cases the improvement is substantial. On the JHMDB dataset, we outperform the method in~\citet{cheron2015p} on 12 sequences against the 21 actions in the dataset. As seen from Table~\ref{tab:mpii_per_class_accuracy}, actions such as \emph{Dry},~\emph{Cut apart}, \emph{Cut slices}, etc. that involve subtle motion cues, benefit most from using KCP.  In Table~\ref{tab:comparison_SMAID_table}, we compare MPII cooking activities classes that are most corrected by SMAID images. We see that actions such as \emph{Change temperature} and \emph{spice}, that involve subtle motions, benefit significantly from SMAID images. Qualitative SMAID images from the MPII cooking activities and the JHMDB dataset are provided in Figure~\ref{fig:qual_results} and Figure~\ref{fig:smaid_quals} using a three-channel SMAID, each channel using 7 frames.

\subsection{Experiments on the Kinetics-600 Dataset}
\label{sec:kinetics}
\newchange{The experiments we presented above use relatively smaller datasets, while there are much larger action recognition datasets available now~\cite{monfort2018moments,zisserman2017kinetics,gu2017ava}. To explore the benefits of our proposed approach to such large scale datasets, we now present experiments on the recently introduced Kinetics-600~\cite{zisserman2017kinetics} dataset\footnote{\url{https://deepmind.com/research/open-source/open-source-datasets/kinetics/}}, which is one of the largest action recognition datasets. This dataset consists of about 460K trimmed video sequences, each video 10 seconds long and annotated for one of 600 pre-defined categories. The dataset is split into 430K training and 30K validation sequences. However, as the dataset only provides Youtube web-links and not videos themselves, not all videos could be downloaded. At the time we ran our experiments only about 390K videos for training and 26,615 videos for validation were available. The rest of the sequences were unavailable despite several downloading attempts. We present results using the available clips.}

\begin{figure}[htbp]
\centering
\subfigure[]{\label{fig:timing_bkcp}\includegraphics[width=4cm,trim={0.1cm 7cm 1cm 7.5cm},clip]{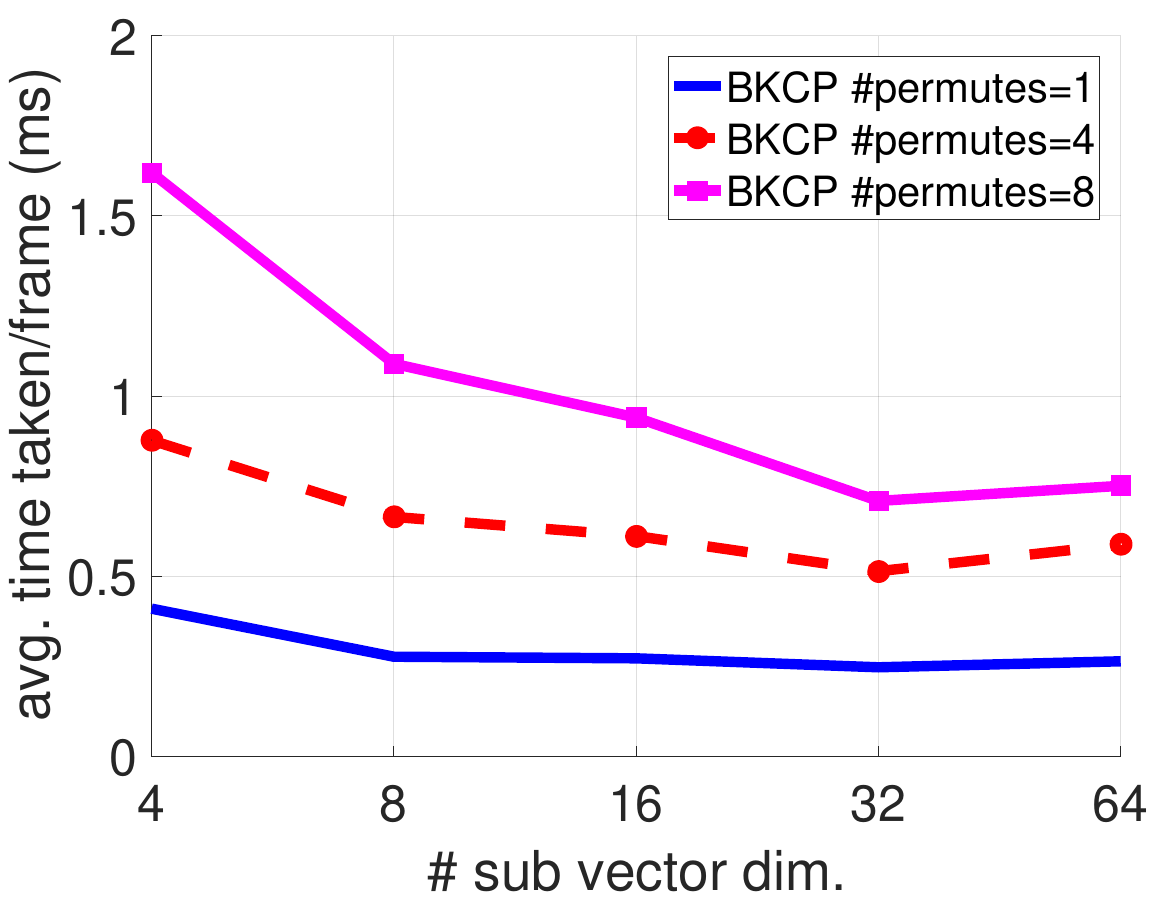}}
\subfigure[]{\label{fig:timing_tcp}\includegraphics[width=4cm,trim={0.1cm 7cm 1cm 7.5cm},clip]{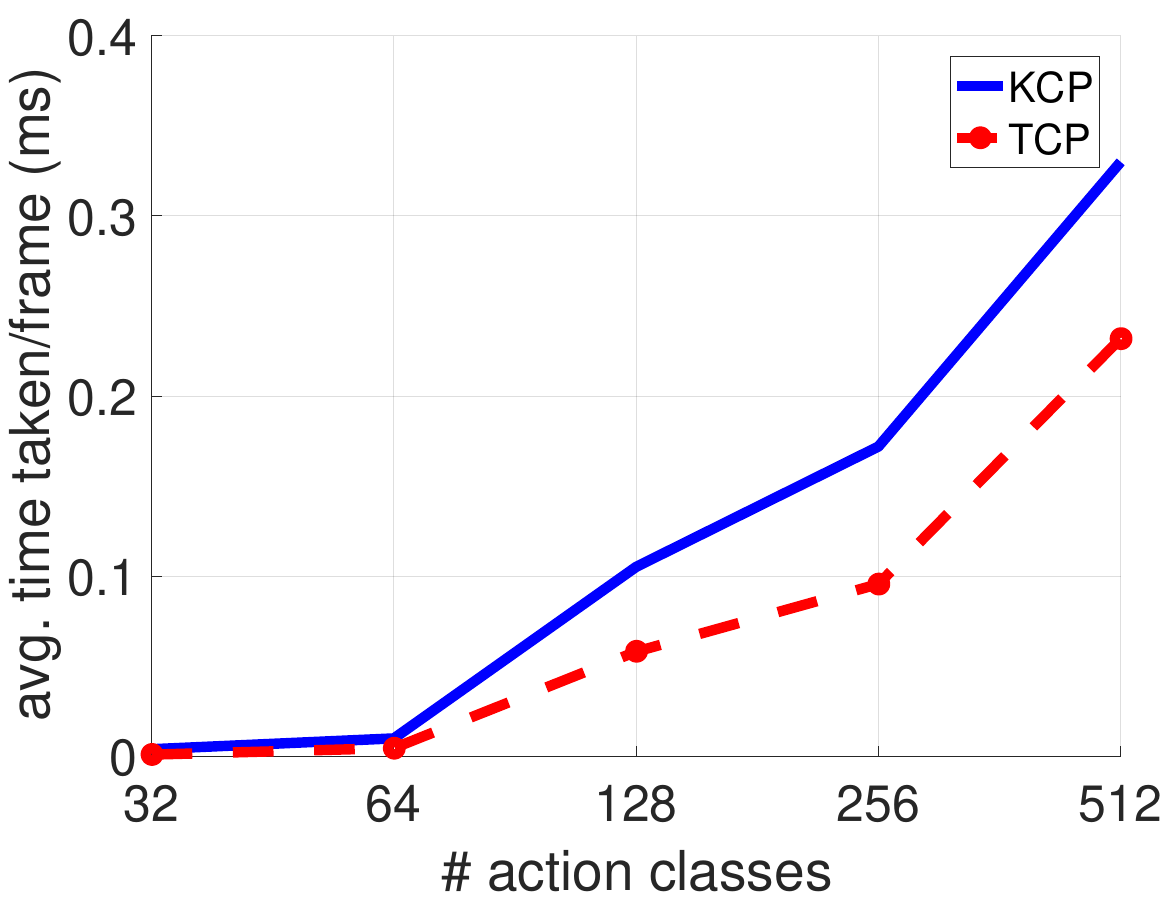}}
\caption{\newchange{Computational performance analysis of BKCP, TCP, and KCP. Left: Time taken against increasing number of permutations and sub-vector dimensionality in BKCP with a feature dimensionality of 4096. Right: Time taken for TCP and BKCP for increasing number of action classes. Plots show average timing in milli-seconds.}}
\label{fig:computations}
\end{figure}

\newchange{For feature extraction, we used the state-of-the-art I3D action recognition model~\cite{carreira2017quo}, which is pre-trained on the earlier Kinetics-400 dataset. As the intention of our experiments is to show the benefits of our pooling scheme on large scale datasets, we did not fine-tune this model on Kinetics-600, instead we passed the RGB sequences through the pre-trained I3D model (using the Inception-V3 architecture), and extracted features from the last pooling layer (Max5c). These features are of size $2\times 7 \times 7 \times 1024$, which are average-pooled across the spatial $7\times 7$ dimensions, and reshaped to 2048-D vectors. We used a sliding window over the clips to generate a sequence of such features with a window size of 16 and a temporal stride of 8 frames. }

\newchange{As we cannot apply our KCP descriptor in the proposed setting (as we do not have a network generating the class confusions), we use the BKCP variant. We explore BKCP with a sub-vector dimensionality of 4 and 8, and the number of permutations  as one or four. The BKCP descriptors are used to train a multi-layer perceptron (MLP) with a single hidden layer of 4096 dimensions. We used the Adam optimizer for training the model with a learning rate of $10^{-4}$ and a batch size of 512. The model was implemented in TensorFlow and the network was trained for about 30K iterations. }

\newchange{The results of this experiment are provided in Table~\ref{tab:kinetics_results}. We compare our BKCP against average pooling (specifically, average pooling all the I3D 2048-D descriptors from a clip, which is technically equivalent to using the original I3D model), followed by training the MLP on these averaged features. The results show that BKCP is much better than using average pooling on the I3D features, and also show that using model averaging over multiple permutations is advantageous, with our best result improving over average pooling by nearly 9\%. We also explored higher values of the sub-vector size and the number of permutations, however the performance was found inferior to the results we report in Table~\ref{tab:kinetics_results}, a trend consistent with Figure~\ref{fig:bkcp-c}. }



\begin{table}
\centering
\begin{tabular}{l|c|c|c|c}
Method & \#perm & p & feat dim. & Accuracy (\%)\\
\hline
I3D~\cite{carreira2017quo} & 1 & 1 & 2048 & 45.9\\
BKCP & 1 & 4 & 3072 & 49.02\\
BKCP & 4 & 4 & 3072 & 50.9\\
BKCP & 1 & 8 & 7168 & 49.1\\
BKCP & 4 & 8 & 7168 & \textbf{54.7}\\
BKCP & 8 & 8 & 7168 & 49.8\\
\end{tabular}
\caption{\newchange{Comparisons on the Kinetics-600 dataset using only the RGB stream. $p$ denotes the sub-vector dimensionality.}}
\label{tab:kinetics_results}
\end{table}

\begin{table}[htbp]
\centering
\begin{tabular}{c|c|c|c|c|c}
RP~\cite{fernando2015modeling} & GRP~\cite{cherian_grp} & BKRP~\cite{cherian2018non} & KRPFS\cite{cherian2018non} & BKCP & KCP\\
\hline
1.1 & 3.8 & 6.7 & 9.5 & 0.43 & 0.005 
\end{tabular}
\caption{\newchange{Avg. run time (time taken / frame) -- in milli-seconds using features of 2048D. BKCP used a sub-vector size of 8 and 4 permutations. KCP uses 51 action classes (as in HMDB dataset).}}
\label{tab:runtime}
\end{table}

\subsection{Run Time Analysis}
\newchange{
In this section, we analyze the computational performance of our schemes for various parameter choices. We present two plots in Figures~\ref{fig:timing_bkcp} and~\ref{fig:timing_tcp}; the former analyzing the time taken by BKCP for various choices of the sub-vector dimension (using a fixed feature dimensionality of 4096) and the number of permutations used in the model averaging, while the latter plotting the performance of TCP and KCP against increasing number of action classes. All the timing experiments used a single core Intel Xeon 2.1GHz CPU and using a Matlab implementation of the algorithms. The timings (in ms) are obtained by averaging over 50 runs. As is clear from the plots, the more the number of permutations used in BKCP, the time taken increases, however is still around 1ms per frame. Further, appropriate choices of the sub-vector dimensionality could improve the effective computational complexity (as we saw in Section~\ref{sec:complexity}. As for the plots in Figure~\ref{fig:timing_tcp}, it is clear that the time taken increases with the number of action classes, and TCP is slightly cheaper than KCP.} 

\newchange{
In Table.~\ref{tab:runtime}, we compare the run time of our descriptors against some recently introduced  pooling schemes, such as rank pooling~\cite{fernando2015modeling}, generalized rank pooling~\cite{cherian_grp}, and the kernelized rank pooling~\cite{cherian2018non}. All these pooling schemes need to solve an optimization scheme for generating the final descriptor. As is clear, our second-order pooling schemes show significantly better  run time performance against these recent schemes, and KCP was found to be 100x faster than for example rank pooling, under the same experimental settings. For all the schemes, we used the Matlab implementations and was run under exactly the same settings.}

\comment{
\begin{table*}[]
	\centering
	\begin{tabular}{c|c|c|c|c|c|c|c}
		Experiment & \small{MPII} & \small{MPII}&  \small{JHMDB} & \small{JHMDB} \\
		&   \small{CP mAP (\%)} & \small{KCP mAP (\%)}  &  \small{CP Avg. Acc. (\%)} & \small{KCP Avg. Acc(\%)}\\
		\hline
		RGB                  & 49.7 & 52.7 &   47.6  & 52.3 &  \\
		FLOW                 & 55.6 & 60.6 &   57.9  & 60.4 \\
		SMAID                  & 51.3 & 55.7 &   50.2  & 64.6 \\
		RGB + FLOW             & 60.0 & 64.4 &   59.4  & 63.4 \\
		RGB + FLOW + SMAID        & 62.1 & 66.1 &   62.0  & 72.7 \\
	\end{tabular}
	\caption{Comparison of CP and KCP on MPII and JHMDB datasets.}
	\label{tab:mpii_jhmdb_cp_kcp_comparison}
\end{table*}

\begin{table}
	\centering
	\begin{tabular}{c|c|c|c|c}
		Expt & \small{MPII} & \small{JHMDB} & \small{HMDB} & \small{UCF101}\\
		\hline
		RGB  &   52.7       &     52.3      &    56.7  &     82.2\\
		FLOW &   60.6       &     60.4      &    53.3 &      85.1\\
		SMAID&   51.3       &     55.7      &    52.9 &      74.2\\
   RGB + FLOW&   64.4       &     63.4      &    65.2 &      91.6\\
RGB + FLOW + SMAID & 66.1   &     72.7      &    66.7 &      90.9\\        
	\end{tabular}
	\caption{Evaluation of KCP between MPII, JHMDB, HMDB(split1) and UCF101 (split1) datasets.}
	\label{tab: kcp_all_datasets.}
\end{table}
}

\comment{
\paragraph{Kernels on the SPD Riemannian manifold:} Next, we evaluate the performance of the different kernels for classifying the correlation matrices. To this end, we use the log-euclidean kernel and the Stein kernel.
We present results using kernelized variants of our correlation pooling scheme. We investigate results using two kernels (i) the Log-Euclidean kernel, and (ii) the Stein kernel. In Table~\ref{tab:kernel_comparison}, we show this comparison, from which it appears that the LE kernel is substantially better on the JHMDB dataset. Further, as this kernel works directly with vectorial data, to which data augmentations (such as including trajectory features is easy), we will use this kernel for further comparisons.
}
\comment{
\begin{table}[]
    \centering
    \begin{tabular}{c|c|c|c|c}
    Experiment & \small{MPII}& \small{JHMDB} \\
                   & \small{KCP mAP(\%)} & \small{KCP Avg.Acc.(\%)}\\
    \hline                   
         LE Kernel        & 66.1 &  72.7 \\
         Stein kernel  & 68.5  & 62.5 \\
    \end{tabular}
    \caption{Comparison of performance when using different kernels in SVM for classifying the kernelized correlation matrices over the three input modalities.}
    \label{tab:kernel_comparison}
\end{table}
}
\comment{
\paragraph{Comparison to Alternative CNN Feature Representations:}
In literature, CNNs have been used as feature generators; these features are extracted from intermediate layers and used to train classifiers. Our goal in this section is to understand the nature of these features and compare it against the representational power of our pooling scheme. To this end, we compare against to such alternatives, namely (i) linear SVMs trained on CNN features extracted from fc6 and fc7 layers after max-pooling and (ii) concatenating the sequence level max-pooled predictions and then training them using an RBF SVM. The results of this experiment are provided in Table~\ref{tab:alternative_classifier_training_comparisons}. We also show the results using our kernelized scheme, which demonstrates superior performance to these alternatives.
}
\comment{
\begin{table}[]
    \centering
    \begin{tabular}{c|c}
        Experiment & \small{MPII-mAP(\%)} \\
        \hline
        max pool of (fc6,fc7) for (Io,Im,Is) + Linear kernel & 64.2\\
        concat(max(Im), max(Io), max(Is)) + RBF kernel    &  62.2 \\
        Kernelized CP + RBF kernel & 66.1\\
    \end{tabular}
    \caption{Comparison of alternative feature representations that could be used to train a non-linear SVM classifier at the sequence level. See text for details.}
    \label{tab:alternative_classifier_training_comparisons}
\end{table}
}
%

%

\comment{
\begin{table}[]
	\centering
	\begin{tabular}{c|c}
		Algorithm &  Avg. Acc. (\%) \\
		\hline
		Two stream~\cite{simonyan2014two} & 88.0\\
		Two stream~\cite{feichtenhofer2016convolutional} & \textbf{93.5}\\
		TDD + IDT~\cite{wang2015action} & 91.5\\
		C3D + IDT~\cite{tran2014c3d} & 90.4\\
		IDT + FV~\cite{wang2013action} & 87.9 \\
		\hline
		KCP & 89.1 \\ 
		BKCP & 88.6 \\
		KCP + BKCP & 89.4\\
		KCP + IDT-FV & 92.0 \\
		BKCP + IDT-FV & 89.3\\
		BKCP + KCP + IDT-FV & 92.4\\
	\end{tabular}
	\caption{UCF101 Dataset (3 splits)}
	\label{tab:ucf101_soa}    
\end{table}
}
\comment{
\begin{table}[]
    \centering
    \begin{tabular}{c|c}
         Algorithm &  Avg. Accuracy (\%) \\
         \hline
         P-CNN~\cite{cheron2015p} & 61.1 \\
         P-CNN + IDT-FV~\cite{cheron2015p} & 72.2 \\
         Action Tubes~\cite{gkioxari2015finding} & 62.5\\
         Stacked Fisher Vectors~\cite{peng2014action} & 69.03\\
         IDT + FV~\cite{wang2013action} & 62.8 \\
         Ground truth Pose + BoW~\cite{jhuang2013towards} & 76.0\\
         Ours (kernelized CP) & \textbf{72.7}\\
         Ours + IDT-FV & \textbf{74.1}\\
    \end{tabular}
    \label{tab:soa_jhmdb}
\end{table}
}

\comment{
	\begin{table}
		\centering
		\begin{tabular}{c|c}
			Algorithm &  Avg. Acc. (\%) \\
			\hline
			Two Stream Fusion, CVPR'16 & 58.17\\
			\hline
			Ours (kernelized CP) & \textbf{58.6}\\
		\end{tabular}
		\caption{HMBD Dataset (split 1)}
		\label{tab:hmdb_soa}
	\end{table}
	}
\comment{\subsection{Results on MPII Cooking Activities Dataset}
\begin{table}[]
    \centering
    \begin{tabular}{c|c|c}
    Expt     &  mAP  & Accuracy\\
    \hline
    softmax of only imflow & 37.6 & 45.3 \\
    softmax of only imsumdiff & 35.3 & 45.4 \\
    softmax of only imdata & 33.9 & 47.0 \\
    Max of softmax imdata + imflow  & 38.1 & 47.8\\
    Max of softmax imdata + imflow + imsumdiff & 39.5 & 49.9\\
    Stacked max pool softmax of imdata + imflow + imsumdiff (RBF) & 62.2   & 63.9 \\
    Stack softmax and then kernelized pool (195x195) & 68.5 & 65.3 \\
    max pool of fc6+fc7 features imdata+ imflow + imsumdiff + traj (linear)    & 64.2 & 67.3 \\
    \hline
    second order pooling of local dynamics (1) & 51.3 & 49.3 \\
    second order pooling of imdata   & 49.7 & 52.0\\
    second order pooling of imflow   & 55.6 & 48.6\\
    kernelized second order pooling of local dynamics & 55.7 & 56.4 \\
    kernelized second order pooling of imdata & 52.5& 55.7 \\
    kernelized second order pooling of imflow   & 60.6 & 53.4\\
    \hline
    second order pooling of flow + imdata + dynamics + trajectories   & 67.8 & 65.8\\
    second order pooling of flow + imdata  + trajectories   & 66.8 &  64.9 \\
    second order pooling of flow + imdata (VGG Two-Stream)   & 60.0 & 58.9   \\
    kernelized second order pooling of flow + imdata (VGG Two-Stream) &  64.4  & 63.7\\
    kernelized second order pooling of flow + imdata + dynamics& 65.6 & 65.4\\
    kernelized second order pooling of flow + imdata + trajectories & 69.8 & 67.9  \\
    kernelized second order pooling of all features + trajectories & 73.5 & 70.0 \\
    Kernelized second order pooling imdata + imflow + sumdiff + stein & 68.5 & 66.4\\
    kernelized second order pooling + all features + Alexnet & \\
    \end{tabular}
    \caption{MPII dataset Ablative study of the influence of various aspects of the algorithm.}
    \label{tab:my_label}
\end{table}
}

\comment{
\begin{table}[]
    \centering
    \begin{tabular}{c|c}
    Expt     &  Mean. Accuracy (\%) \\
    \hline
    max pooling of softmax features         & \\
    \hline
    second order pooling of local dynamics  & 50.2 \\
    second order pooling of imflow & 57.9 \\
    second order pooling of imdata & 47.6 \\
    kernelized second order pooling of local dynamics & 64.6\\
    kernelized second order pooling of imdata  & 51.5 \\
    kernelized second order pooling of imflow  & 60.4\\
    \hline
    second order pooling of flow + imdata (VGG Two-Stream)   & 59.4    \\
    second order pooling of flow + imdata + dynamics   &   62.0  \\
    kernelized second order pooling of flow + imdata (VGG Two-Stream) &  63.4 \\
    kernelized second order pooling of flow + imdata + dynamics & 72.7\\
    \end{tabular}
    \caption{JHMDB dataset influence of various aspects of the algorithm.}
    \label{tab:my_label}
\end{table}
}

%% file: conclusions.tex
\section{Conclusions}
\label{sec:conclude}
In this paper, we proposed a temporal pooling scheme,~\emph{temporal correlation pooling}, based on the correlations between temporal evolution of classifier scores. Our scheme produces positive definite matrices as descriptors, allowing the use rich mathematical (Riemannian) geometries for non-linear feature pooling. While, our basic descriptor scales quadratically against the number of action classes, we proposed a simple approximation to it that scales linearly. We also proposed a novel sub-sequence representation, SMAID, that can dramatically increase the temporal receptive fields of CNNs, thereby improving action classification performance. Using SMAID and temporal correlation pooling schemes, we proposed a novel three-stream end-to-end learnable CNN architecture for action classification. The utility of each of our contributions was substantiated via experiments on four challenging action recognition benchmarks, and the scalability and relevance of the scheme further substantiated via experiments on the Kinetics-600 action recognition dataset.

\comment{
Our representation is computationally cheap, and can be learned in an end-to-end CNN learning framework. 
Next, frame-level classifiers trained on this representation are more accuratewe also provided extensions of this scheme using kernel representations and block-diagonal approximations to handle high-dimensionality of data features. Our experiments clearly demonstrated the advantages of the proposed schemes, leading to state-of-the-art results on four benchmark datasets. 
}